\RequirePackage{fix-cm}
\documentclass[a4paper,11pt]{article}

\usepackage{authblk}
\usepackage{fancyhdr}
\usepackage{amssymb}
\usepackage{multirow}
\usepackage{graphicx}
\usepackage[caption=false]{subfig}
\usepackage{amsmath}
\usepackage{amssymb}
\usepackage{amsfonts}
\usepackage{cite}
\usepackage{wasysym}
\usepackage{microtype}
\usepackage{todonotes}
\usepackage{tikz}
\usepackage{hyperref}
\usepackage{booktabs}
\usepackage{paralist}
\usetikzlibrary{positioning}
\usetikzlibrary{decorations.pathreplacing}

\tikzset{
	mybrace/.style={decorate,decoration={brace,aspect=#1}}
}

\newcommand{\N}{\mathbb{N}}
\newcommand{\Z}{\mathbb{Z}}
\newcommand{\F}{\mathbb{F}}

\newtheorem{definition}{Definition}
\newtheorem{lemma}{Lemma}

\newtheorem{problem}{Problem}
\newtheorem{example}{Example}

\providecommand{\keywords}[1]{\textbf{\textit{Keywords }} #1}

\begin{document}

\title{Evolutionary Algorithms for Designing \\ Reversible Cellular Automata}

\author[1]{Luca Mariot}
\author[1]{Stjepan Picek}
\author[2]{Domagoj Jakobovic}
\author[3]{Alberto Leporati}

\affil[1]{{\normalsize Cyber Security Research Group, Delft University of Technology, Mekelweg 2, Delft, The Netherlands} \\
	
	{\small \texttt{\{l.mariot, s.picek\}@tudelft.nl}}}

\affil[2]{{\normalsize Faculty of Electrical Engineering and Computing, University of Zagreb, Unska 3, Zagreb, Croatia} \\
	
	{\small \texttt{domagoj.jakobovic@fer.hr}}}

\affil[3]{{\normalsize DISCo, Università degli Studi di Milano-Bicocca, Viale Sarca 336/14, Milano, Italy} \\
	
	{\small \texttt{alberto.leporati@unimib.it}}}

\maketitle

\begin{abstract}
  Reversible Cellular Automata (RCA) are a particular kind of shift-invariant transformations characterized by a dynamics composed only of disjoint cycles. They have many applications in the simulation of physical systems, cryptography and reversible computing. In this work, we formulate the search of a specific class of RCA  -- namely, those whose local update rules are defined by conserved landscapes -- as an optimization problem to be tackled with Genetic Algorithms (GA) and Genetic Programming (GP). In particular, our experimental investigation revolves around three different research questions, which we address through a single-objective, a multi-objective, and a lexicographic approach. The results obtained from our experiments corroborate the previous findings and shed new light on 1) the difficulty of the associated optimization problem for GA and GP, 2) the relevance of conserved landscape CA in the domain of cryptography and reversible computing, and 3) the relationship between the reversibility property and the Hamming weight.
\end{abstract}


\section{Introduction}
\label{sec:intro}

The shift-invariance property is important when studying and modeling several types of discrete dynamical systems.
The property states that any translation of the input state results in the same translation of the output state in a system governed by a shift-invariant transformation. When a finite array describes the state of the system, shift-invariant transformations are cellular automata (CA), i.e., functions defined by a local update rule uniformly applied at all sites of the array. CA have been thoroughly studied both as models for simulating discrete dynamical systems in physics~\cite{toffoli90}, biology~\cite{green90,sirakoulis03}, ecology~\cite{hogeweg88,mariot18} and other fields, as well as to design computational devices, for example in symmetric cryptography~\cite{daemen95,mariot19} and fault-tolerant computing~\cite{nishio75}.
Reversible shift-invariant transformations, particularly Reversible CA (RCA), have the additional characteristic of preserving information. As such, the dynamics of an RCA can be reversed backward in time starting from any state, and the inverse mapping is itself a CA. This characteristic makes RCA especially interesting for designing energy-efficient computing devices, as stated by Landauer’s principle~\cite{landauer61}. In fact, any irreversible logical operation implemented in hardware leads to the dissipation of heat, and this entails a physical lower bound on the miniaturization of devices based on irreversible gates. One more interesting domain for RCA is cryptography, where they can be used to design encryption and decryption algorithms~\cite{mariot19}.

Unfortunately, while RCA are characterized by simple combinatorial rules, designing them is a difficult problem when considering additional properties as required by specific applications. This is because there are only a few known classes of RCA~\cite{kari18} and an exhaustive search of all possible RCA is unfeasible for large local rule sizes. 
Considering these difficulties and the limited number of available theoretical results, heuristics -- and, more precisely, evolutionary algorithms (EA) -- represent an interesting option for designing RCA.

An interesting class of CA that include reversible ones are \emph{marker CA}, where the local update rule flips the state of a cell if its neighbors take on a set of patterns (also called \emph{flipping landscapes}) that are conserved by the resulting shift-invariant transformation~\cite{toffoli90}. 
Evolutionary algorithms like genetic algorithms (GA) and genetic programming (GP) intuitively represent a good fit to evolve the local rules of marker CA since they have a simple description through their generating functions. In particular, the output of a marker CA rule corresponds to the XOR of the cell in the origin of the neighborhood and its generating function evaluated on the neighboring cells. As such, it becomes rather straightforward to formulate the optimization objective for the reversibility property by minimizing the number of compatible flipping landscapes defined by the generating function. An optimal solution is a marker CA rule whose flipping landscapes are mutually incompatible, or equivalently a \emph{conserved landscape rule}.

Additionally, the Hamming weight of a generating function in a marker CA represents a good indicator of its 1) \emph{nonlinearity}, which is a relevant property in domains like 
sequences~\cite{1056589}, telecommunication~\cite{Paterson}, and cryptography~\cite{mariot19}. Consequently, maximizing the Hamming weight of the generating function can be considered as an additional optimization objective.

Our research investigates how difficult it is for evolutionary algorithms to find conserved landscape CA rules, considering their small number as compared to the corresponding search space size. Additionally, we explore the evolution of rules of larger diameter (i.e., larger neighborhood size), as such rules are relevant from the practical perspective. Finally, we investigate the trade-offs between the reversibility of a marker CA rule and the Hamming weight.

This paper is an extended version of the work ``An Evolutionary View on Reversible
Shift-Invariant Transformations''~\cite{10.1007/978-3-030-44094-7_8} presented at EuroGP 2020. With respect to that work, here:
\begin{compactenum}
	\item We consider one additional evolutionary algorithm in our experiments, namely the lexicographic genetic algorithm. By doing so, we allow a more detailed analysis of the lexicographic paradigm for the evolution of conserved landscape CA.
	\item We conduct an extensive tuning phase for all algorithms on the problem instance with diameter $d$ equal to 10. This represents a much larger and more difficult problem than the one considered in~\cite{10.1007/978-3-030-44094-7_8}, where the diameter for tuning was set to 7, thus allowing more meaningful tuning results. 
	\item We consider more problem instances: while the original paper considered diameter sizes $d$ from 8 to 13, this work investigates diameter sizes ranging from 7 to 15. 
	\item While in the original paper we allowed the offset $\omega$ to be of size $d-1$, here we set $\omega$ equal to 3 for all experiments. By doing so, we aim to explore a more difficult optimization problem, as there will be fewer solutions fulfilling the criteria.
	\item Finally, we provide a more detailed experimental analysis by also considering Hamming weight distributions and algorithms' convergence.
\end{compactenum}
Besides confirming the observations from~\cite{10.1007/978-3-030-44094-7_8}, the new set of experiments allowed us to discover two additional findings:
\begin{compactitem}
	\item We show that GP manages to find optimal solutions already in the initial population. This indicates that although decreasing $\omega$ limits the total number of optimal solutions, it still allows GP to ``easily'' guess some of those solutions. Thus, having smaller $\omega$ makes the problem simpler for GP, but not for GA, where we observed a trend of increasing difficulty similar to the one reported in~\cite{10.1007/978-3-030-44094-7_8}. 
	\item The Pareto fronts obtained with the multi-objective optimization approach indicate not only that the Hamming weight of an optimal solution must necessarily be low concerning the length of its truth table, but also that balanced generating functions are the farthest possible from giving reversible rules. This, in turn, allows us to further explain why for GA, it is extremely unlikely to guess an optimal solution by chance in the initial population, while it is easy for GP.
\end{compactitem}

The rest of this paper is organized as follows. In Section~\ref{sec:ca}, we discuss some types of cellular automata, and we provide some relevant definitions and notations. Section~\ref{sec:rel-works} presents related works. In Section~\ref{sec:opt-rev}, we discuss how to optimize the reversibility of CA, presenting a first attempt based on the de Bruijn graph representation of a local rule and then discarding it in favor of the approach based on conserved landscape rules. Section~\ref{sec:exp} first presents the result of a preliminary exhaustive search and then provides details on our experimental setting and parameter tuning phase. In Section~\ref{sec:results} we present the results of our evolutionary experiments, while in Section~\ref{sec:disc} we discuss them with respect to the stated research questions. Finally, Section~\ref{sec:conclusions} concludes the paper and offers potential directions for future research.

\section{Cellular Automata (CA)}
\label{sec:ca}

This section covers background definitions and notions on reversible cellular automata, upon which the rest of the paper is based. We start with some general definitions, followed by discussions on reversible CA and marker CA.

\subsection{Basic Definitions}
\label{subsec:bas-def}

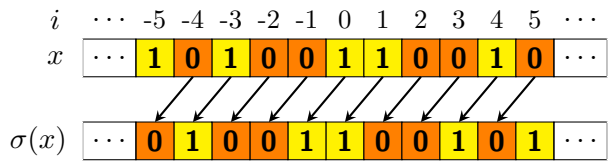
\begin{figure}[t]
	\centering
	\begin{tikzpicture}
	[->,auto,node distance=1.5cm, empt node/.style={font=\sffamily,inner
		sep=0pt,minimum size=0.3cm}, 
	rect node/.style={rectangle,draw,font=\sffamily\bfseries,minimum size=0.5cm, inner sep=0pt, outer sep=0pt},
	rect0 node/.style={rectangle,draw,fill=orange,font=\sffamily\bfseries,minimum size=0.5cm, inner sep=0pt, outer sep=0pt},
	rect1 node/.style={rectangle,draw,fill=yellow,font=\sffamily\bfseries,minimum size=0.5cm, inner sep=0pt, outer sep=0pt},
	]
	
	\node [rect1 node] (c0) {1};
	\node [rect1 node] (c1) [right=0cm of c0] {1};
	\node [rect0 node] (c2) [right=0cm of c1] {0};
	\node [rect0 node] (c3) [right=0cm of c2] {0};
	\node [rect1 node] (c4) [right=0cm of c3] {1};
	\node [rect0 node] (c5) [right=0cm of c4] {0};
	\node [rect node] (c6) [right=0cm of c5, minimum width=0.7cm] {$\ldots$};
	\node [rect node] (c7) [right=0cm of c6, draw=white] {};
	\node [rect0 node] (c-1) [left=0cm of c0] {0};
	\node [rect0 node] (c-2) [left=0cm of c-1] {0};
	\node [rect1 node] (c-3) [left=0cm of c-2] {1};
	\node [rect0 node] (c-4) [left=0cm of c-3] {0};
	\node [rect1 node] (c-5) [left=0cm of c-4] {1};
	\node [rect node] (c-6) [left=0cm of c-5, minimum width=0.7cm] {$\ldots$};
	\node [rect node] (c-7) [left=0cm of c-6, draw=white] {};
	\node [empt node] (x) [left=0.2cm of c-6] {$x$};
	\node [empt node] (e0) [above=0.1cm of c0] {\small $0$};
	\node [empt node] (e1) [above=0.1cm of c1] {\small $1$};
	\node [empt node] (e2) [above=0.1cm of c2] {\small $2$};
	\node [empt node] (e3) [above=0.1cm of c3] {\small $3$};
	\node [empt node] (e4) [above=0.1cm of c4] {\small $4$};
	\node [empt node] (e5) [above=0.1cm of c5] {\small $5$};
	\node [empt node] (e6) [above=0.1cm of c6, minimum width=0.7cm] {$\ldots$};
	\node [empt node] (e-1) [above=0.1cm of c-1] {\small -$1$};
	\node [empt node] (e-2) [above=0.1cm of c-2] {\small -$2$};
	\node [empt node] (e-3) [above=0.1cm of c-3] {\small -$3$};
	\node [empt node] (e-4) [above=0.1cm of c-4] {\small -$4$};
	\node [empt node] (e-5) [above=0.1cm of c-5] {\small -$5$};
	\node [empt node] (e-6) [above=0.1cm of c-6, minimum width=0.7cm] {$\ldots$};
	\node [empt node] (i) [left=0.2cm of e-6] {$i$};
	
	\node [rect1 node] (d0) [below=0.6cm of c0] {1};
	\node [rect0 node] (d1) [right=0cm of d0] {0};
	\node [rect0 node] (d2) [right=0cm of d1] {0};
	\node [rect1 node] (d3) [right=0cm of d2] {1};
	\node [rect0 node] (d4) [right=0cm of d3] {0};
	\node [rect1 node] (d5) [right=0cm of d4] {1};
	\node [rect node] (d6) [right=0cm of d5, minimum width=0.7cm] {$\ldots$};
	\node [rect node] (d7) [right=0cm of d6, draw=white] {};
	\node [rect1 node] (d-1) [left=0cm of d0] {1};
	\node [rect0 node] (d-2) [left=0cm of d-1] {0};
	\node [rect0 node] (d-3) [left=0cm of d-2] {0};
	\node [rect1 node] (d-4) [left=0cm of d-3] {1};
	\node [rect0 node] (d-5) [left=0cm of d-4] {0};
	\node [rect node] (d-6) [left=0cm of d-5, minimum width=0.7cm] {$\ldots$};
	\node [rect node] (c-7) [left=0cm of d-6, draw=white] {};
	\node [empt node] (x) [left=0.2cm of d-6] {$\sigma(x)$};
	
	\draw[->,thick,shorten >=0pt,shorten <=0pt,>=stealth] (c5.south) -- (d4.north);
	\draw[->,thick,shorten >=0pt,shorten <=0pt,>=stealth] (c4.south) -- (d3.north);
	\draw[->,thick,shorten >=0pt,shorten <=0pt,>=stealth] (c3.south) -- (d2.north);
	\draw[->,thick,shorten >=0pt,shorten <=0pt,>=stealth] (c2.south) -- (d1.north);
	\draw[->,thick,shorten >=0pt,shorten <=0pt,>=stealth] (c1.south) -- (d0.north);
	\draw[->,thick,shorten >=0pt,shorten <=0pt,>=stealth] (c0.south) -- (d-1.north);
	\draw[->,thick,shorten >=0pt,shorten <=0pt,>=stealth] (c-1.south) -- (d-2.north);
	\draw[->,thick,shorten >=0pt,shorten <=0pt,>=stealth] (c-2.south) -- (d-3.north);
	\draw[->,thick,shorten >=0pt,shorten <=0pt,>=stealth] (c-3.south) -- (d-4.north);
	\draw[->,thick,shorten >=0pt,shorten <=0pt,>=stealth] (c-4.south) -- (d-5.north);
	
	\end{tikzpicture}
	\caption{Example of application of the shift operator $\sigma$ over a bi-infinite string $x$ in $A^{\Z}$. The alphabet in this example is $A=\{0,1\}$.}
	\label{fig:shift}
\end{figure}

Let $A$ be a finite alphabet, and let us denote by $A^{\Z}$ the set of all \emph{bi-infinite strings} over $A$. The \emph{shift operator} is defined as the function $\sigma: A^{\Z} \rightarrow A^{\Z}$ that takes as input a bi-infinite string $x \in A^{\Z}$ and shifts each component of it one place to the left, as shown in Figure~\ref{fig:shift}.

In the field of \emph{symbolic dynamics}~\cite{lind21}, the set $A^{\Z}$ equipped with the shift operator is also called the \emph{full-shift space}. This set can be further endowed with the \emph{Cantor distance}, under which two configurations $x,y \in A^{\Z}$ are close to one another if the first coordinate $i$ where they differ is close to the origin. The topological space resulting from this distance is \emph{compact}~\cite{kurka09}, simplifying the study of mappings $F: A^{\Z} \to A^{\Z}$ as \emph{dynamical systems}. In particular, \emph{shift-invariant transformations} are those mappings $F: A^{\Z} \to A^{\Z}$ that commute with the \emph{shift operator}, that is, those mappings for which
\begin{displaymath}
	F(\sigma(x)) = \sigma(F(x)) \enspace , \enspace \textrm{for all } x \in \{0,1\}^{\Z}.
\end{displaymath}

\emph{Cellular Automata} (CA) are a particular class of shift-invariant transformations whose output is determined by the parallel application of a single \emph{local update rule} over all components (or \emph{cells}) of a bi-infinite string. Such a rule depends only on a finite number of neighboring cells, also called the \emph{diameter}. The \emph{Curtis-Hedlund-Lyndon (CHL) Theorem} characterizes CA as those mappings $F: A^{\Z} \to A^{\Z}$ that are both shift-invariant and uniformly continuous with respect to the Cantor distance~\cite{hedlund69}. When restricting the attention only to the subset of \emph{spatially periodic configurations}, i.e., those configurations $x \in A^{\Z}$ that repeat themselves after a minimum period $p \in \N$, any shift-invariant transformation is described by a CA. Indeed, the continuity requirement of the CHL theorem can be dropped since the diameter of the CA local rule is upper bounded by the period $p$ of the configurations. Hence, the next state of each cell cannot depend on cells that are arbitrarily far apart. In particular, the case of shift-invariant transformations over finite arrays of length $p \in \N$ with \emph{periodic boundary conditions} (i.e., where the array can be seen as a ``ring'' in which the first cell follows the last one) coincides with CA over the set of spatially periodic configurations $x \in A^\Z$ having period $p$. Clearly, finite CA (or equivalently, infinite CA over periodic configurations) represent the most interesting case for practical applications, and we focus exclusively on them in the rest of this paper. Therefore, in what follows, we use the term CA and shift-invariant transformation interchangeably.

Various CA models can be defined depending on the dimension of the cellular array, the alphabet of the cells, and the boundary conditions. In this work, we focus on \emph{one-dimensional} \emph{periodic} \emph{Boolean} CA, defined as follows:
\begin{definition}
	\label{def:ca}
	A one-dimensional periodic Boolean CA (PBCA) of length $n$, diameter $d$, offset $\omega$, and local rule $f: \{0,1\}^d \rightarrow \{0,1\}$, is defined as a vectorial function $F: \{0,1\}^{n} \rightarrow \{0,1\}^{n}$ where
	for every vector $x \in \{0,1\}^n$ and all $0\le i \le n-1$, the $i$-th component of the output vector is given by:
	\begin{equation}
		\label{eq:grule-pbca}
		F(x)_i = f(x_{[i-\omega,i-\omega+d-1]}) = f(x_{i-\omega}, x_{i-\omega+1},
		\cdots, x_{i}, \cdots, x_{i-\omega+d-1})
	\end{equation}
	with all indices being computed modulo $n$. Function $F$ is also called the \emph{global rule} of the CA.
\end{definition}
Thus, a PBCA is composed of a one-dimensional vector of $n$ cells that can be either in state $0$ or $1$, where each cell simultaneously updates its state by applying the local rule $f$ on the neighborhood formed by itself, the $\omega$ cells on its left and the $d-1-\omega$ cells on its right. Here, ``periodic'' refers to the fact that all indices are computed modulo $n$: in this way, the leftmost $\omega$ cells and the rightmost $d-1-\omega$ ones respectively have enough left and right neighboring cells to apply the local rule. Unless ambiguities arise, in what follows we refer to PBCA simply as CA, as the former is the main CA model considered in this work. The \emph{orbit} of a PBCA starting from $x$ is the sequence of vectors $\{x(t)\}_{t \in \N}$ where $x(0) = x \in \F_2^{n}$ and $x(t) = F^t(x)$ for all $t>0$ (remark that $F^t$ denotes the iteration of the CA global rule $F$ for $t$ times).

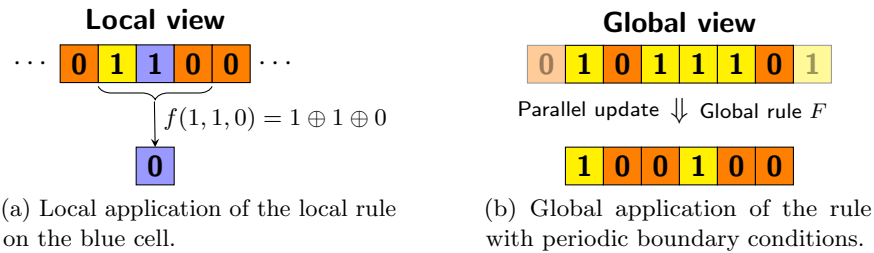
\begin{figure}[t]
	\centering
	\subfloat[Local application of the local rule on the blue cell.\label{fig:ex-150-tab-cl}]{
		\centering
		\begin{tikzpicture}
		[->,auto,node distance=1.5cm, empt node/.style={font=\sffamily,inner
			sep=0pt,minimum size=0pt}, rect0 node/.style={rectangle,draw,fill=orange,font=\sffamily\bfseries,minimum size=0.5cm, inner
			sep=0pt, outer sep=0pt},
		rect1 node/.style={rectangle,draw,fill=yellow,font=\sffamily\bfseries,minimum size=0.5cm, inner
			sep=0pt, outer sep=0pt}, blue node/.style={rectangle,draw,fill=blue!40,
			font=\sffamily\bfseries,minimum size=0.5cm, inner sep=0pt, outer sep=0pt}]
		
		\node [blue node] (c)   {0};
		\node [empt node] (f1) [above=0.2cm of c] {\phantom{$\downarrow$}};
		\node [empt node] (f2) [right=0cm of f1] {{\footnotesize $f(1,1,0) = 1
				\oplus 1 \oplus 0$}};
		\node [empt node] (f3) [left=-0.1cm of f1] {};
		\node [empt node] (f4) [above=0.22cm of f3] {};
		
		\node [blue node] (p) [above=0.3cm of f1] {1};
		\node [empt node] (text) [above=0.2cm of p] {\bfseries Local view};
		\node [rect1 node] (pl1) [left=0cm of p] {1};
		\node [rect0 node] (pl2) [left=0cm of pl1] {0};
		\node [empt node] (pl3) [left=0.1cm of pl2] {$\cdots$};
		\node [rect0 node] (pr1) [right=0cm of p] {0};
		\node [rect0 node] (pr2) [right=0cm of pr1] {0};
		\node [empt node] (pr3) [right=0.1cm of pr2] {$\cdots$};
		
		\draw [-, decorate,
		decoration={brace,mirror,amplitude=5pt,raise=0.3cm}] (pl1.west) --
		(pr1.east) node [midway] {};
		
		\draw [>=stealth] (f4) -- (c.north);
		
		%\draw [-, decorate, decoration={brace,mirror,amplitude=5pt,raise=0.3cm}]
		%(pl1.west) -- (pr1.east) node [midway,yshift=-0.2cm] {};
		
		\end{tikzpicture}
	}%
	\phantom{MMM}
	\subfloat[Global application of the rule with periodic boundary conditions.\label{fig:ex-150-tab-cr}]{
		\centering
		\begin{tikzpicture}
		[->,auto,node distance=1.5cm, empt node/.style={font=\sffamily,inner
			sep=0pt,minimum size=0.3cm}, rect0 node/.style={rectangle,draw,fill=orange,font=\sffamily\bfseries,minimum size=0.5cm, inner
			sep=0pt, outer sep=0pt},
		rect1 node/.style={rectangle,draw,fill=yellow,font=\sffamily\bfseries,minimum size=0.5cm, inner
			sep=0pt, outer sep=0pt}]
		
		\node [empt node] (c)   {};
		\node [rect1 node] (c1) [right=0.1cm of c] {1};
		\node [rect0 node] (c2) [right=0cm of c1] {0};
		\node [rect0 node] (c3) [right=0cm of c2] {0};
		\node [rect1 node] (c4) [right=0cm of c3] {1};
		\node [rect0 node] (c5) [right=0cm of c4] {0};
		\node [rect0 node] (c6) [right=0cm of c5] {0};
		\node [empt node] (c7) [right=0.1cm of c6] {};
		
		\node [empt node] (f1) [above=0.55cm of c3.east] {$\Downarrow$};
		\node [empt node] (f2) [left=0.1cm of f1] {{\scriptsize Parallel update}};
		\node [empt node] (f3) [right=0.1cm of f1] {{\scriptsize Global rule $F$}};
		\node [rect1 node] (p2) [above=0.85cm of c1] {1};
		\node [rect0 node] (p1) [left=0cm of p2, opacity=0.4] {0};
		\node [empt node] (p)  [left=0.1cm of p1] {};
		\node [rect0 node] (p3) [right=0cm of p2] {0};
		\node [rect1 node] (p4) [right=0cm of p3] {1};
		\node [empt node] (text1) [above=0.4cm of p4.east] {\bfseries Global view};
		\node [rect1 node] (p5) [right=0cm of p4] {1};
		\node [rect1 node] (p6) [right=0cm of p5] {1};
		\node [rect0 node] (p7) [right=0cm of p6] {0};
		\node [rect1 node] (p8) [right=0cm of p7, opacity=0.4] {1};
		\node [empt node] (p9) [right=0.1cm of p8] {};
		\end{tikzpicture}
	}
	\caption{An example of CA with $n = 6$ cells, diameter $d = 3$, offset $\omega = 1$, and local rule defined as $f(x_{i-1},x_i,x_{i+1}) = x_{i-1} \oplus x_i \oplus x_{i+1}$, corresponding to Wolfram code 150.}
	\label{fig:ex-150}
\end{figure}

Since the cells of a CA take binary values, the local rule can be seen as a \emph{Boolean function} $f: \F_2^d \rightarrow \F_2$ \ of $d$ variables where $\F_2 = \{0,1\}$ is the finite field of two elements, and thus it can be represented by its \emph{truth table}, which specifies for each of the possible $2^d$ input vectors $x \in \F_2^d$ the corresponding output value $f(x) \in \F_2$.
Assuming that the input vectors of $\F_2^d$ are sorted lexicographically (i.e., $x\le y$ if and only if $x_i \le y_i$ where $i$ is the first index such that $x_i$ and $y_i$ differ), one can encode the truth table as a single binary string $\Omega_f \in \F_2^{2^d}$, which is the output column of the table. In the CA literature, the decimal encoding of $\Omega_f$ is also called the \emph{Wolfram code} of the local rule $f$~\cite{wolfram83}. Figure~\ref{fig:ex-150} reports an example of CA with $n=6$ cells, diameter $d=3$, offset $\omega = 1$, and local rule defined as $f(x_{i-1},x_{i},x_{i+1}) = x_{i-1}\oplus x_{i} \oplus x_{i+1}$, corresponding to Wolfram code 150.
Hence, each cell looks at itself and at its left and right neighbors to compute its next state through rule 150. The two shaded cells in Figure~\ref{fig:ex-150-tab-cr} represent ``copies'' respectively of the first and of the last cell, to help visualizing the neighborhoods of the cells at the boundaries. As mentioned above, one can effectively think of the CA array as a ring, bending it so that the leftmost and rightmost cells come close to each other.

Besides the truth table, another useful way to represent the local rule of a CA is by means of a \emph{de Bruijn graph}, as shown originally by Sutner~\cite{sutner91}. Such a representation is based on the \emph{overlap operator} $\odot$ that, given two vectors $x,y \in \F_2^{d-1}$ with $x = (x_1,x_2,\cdots, x_{d-1})$ and $y = (y_1, y_2, \cdots, y_{d-1})$ such that $x_i = y_{i-1}$ for all $2 \le i \le d-1$, outputs the vector $x \odot y = z \in \F_2^d$ defined as $z = (x_1, x_2, \cdots, x_{d-1}, y_{d-1})$. In other words, $z$ is obtained by \emph{overlapping} the rightmost $d-2$ components of $x$ with the leftmost $d-2$ ones of $y$, thereby obtaining a vector of length $d$.
Formally, the de Bruijn graph of a CA with local rule $f: \F_2^d \to \F_2$ is defined as:
\begin{definition}
	\label{def:db-graph}
	Let $F: \F_2^n \to \F_2^n$ be a CA equipped with a local rule $f: \F_2^{d} \rightarrow \F_2$ of diameter $d\le n$. Then, the \emph{de Bruijn graph} associated to $F$ is the directed labeled graph $G_{DB}(f) = (V,E,l)$ where $V = \F_2^{d-1}$. Further, given $v_1, v_2 \in V$, it holds that $(v_1,v_2) \in E$ if and only if there exists $z \in \F_2^{d}$ such that $z = v_1 \odot v_2$, i.e., if $v_1$ and $v_2$ can be overlapped. Finally, for all $(v_1, v_2) \in E$, the label function $l: E \rightarrow \F_2$ is defined as $l(v_1,v_2) = f(v_1 \odot v_2)$.
\end{definition}

\begin{figure}[t]
	\centering
	\subfloat[Truth table.\label{fig:tt-150}]{
		\centering
		\raisebox{1.45cm}{
			\begin{tabular}{ccc|c}
				\hline
				$x_1$ & $x_2$ & $x_3$ & $f(x_1,x_2,x_3)$ \\
				\hline
				$0$   & $0$  & $0$   &  $0$ \\
				$1$   & $0$  & $0$   &  $1$ \\
				$0$   & $1$  & $0$   &  $1$ \\
				$1$   & $1$  & $0$   &  $0$ \\
				$0$   & $0$  & $1$   &  $1$ \\
				$1$   & $0$  & $1$   &  $0$ \\
				$0$   & $1$  & $1$   &  $0$ \\
				$1$   & $1$  & $1$   &  $1$ \\
				\hline
			\end{tabular}
		}
	}%
	\phantom{MMMM}
	\subfloat[De Bruijn Graph.\label{fig:db-150}]{
		\centering
		\Large
		\resizebox{3cm}{!}{
			\begin{tikzpicture}
			[->,auto,node distance=0.5cm, every loop/.style={min distance=10mm},
			empt node/.style={font=\sffamily,inner sep=0pt,outer sep=0pt},
			circ node/.style={circle,thick,draw,font=\sffamily\bfseries,minimum
				width=0.8cm, inner sep=0pt, outer sep=0pt}]
			
			% Nodes
			\node [empt node] (e1) {};
			\node [circ node] (n00) [above=1.75cm of e1] {$00$};
			\node [circ node] (n01) [right=1.75cm of e1] {$01$};
			\node [circ node] (n10) [left=1.75cm of e1] {$10$};
			\node [circ node] (n11) [below=1.75cm of e1] {$11$};       
			
			% Edges
			\draw [->, thick, shorten >=0pt,shorten <=0pt,>=stealth] (n00) 
			edge[bend left=20] node (f5) [above right]{$1$} (n01);
			\draw [->, thick, shorten >=0pt,shorten <=0pt,>=stealth] (n01)
			edge[bend left=20] node (f5) [below right]{$0$} (n11);
			\draw [->, thick, shorten >=0pt,shorten <=0pt,>=stealth] (n11)
			edge[bend left=20] node (f5) [below left]{$0$} (n10);
			\draw [->, thick, shorten >=0pt,shorten <=0pt,>=stealth] (n10)
			edge[bend left=20] node (f5) [above left]{$1$} (n00);
			\draw[->, thick, shorten >=0pt,shorten <=0pt,>=stealth] (n10) edge[bend
			left=20] node (f1) [above]{$0$} (n01);
			\draw[->, thick, shorten >=0pt,shorten <=0pt,>=stealth] (n01)
			edge[bend left=20] node (f2) [below]{$1$} (n10);
			\draw[->, thick, shorten >=0pt,shorten <=0pt,>=stealth] (n00) edge[loop
			above] node (f3) [above]{$0$} ();
			\draw[->, thick, shorten >=0pt,shorten <=0pt,>=stealth] (n11) edge[loop
			below] node (f4) [below]{$1$} ();
			\end{tikzpicture}
		}
	}
	\caption{Truth table and de Bruijn graph representations for local rule 150.}
	\label{fig:ex-tt-db-150}
\end{figure}
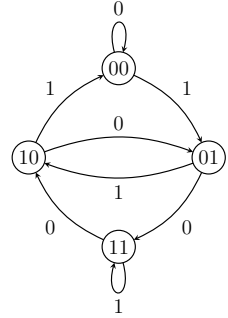

As an example, Figure~\ref{fig:ex-tt-db-150} reports the truth table and the de Bruijn graph representations associated to the CA $F$ with local rule $150$.
In particular, the input of a CA $F: \F_2^n \to \F_2^n$ is a path of length $n-d+1$ \emph{on the vertices} of its de Bruijn graph, obtained by overlapping all vertices in the path to get a vector $x \in \F_2^n$, while $F(x)$ is the corresponding path \emph{on the edges}, obtained by concatenating the labels.

\subsection{Reversible CA}
\label{subsec:rev-ca}

The property of reversibility is of particular importance in the field of dynamical systems. Stated informally, the orbits of the states of a reversible system are disjoint cycles without transient parts or pre-periods. Consequently, the dynamics of such systems can also be run backward in the time since each state has exactly one predecessor, and the inverse system is analogous to the original one. In the context of infinite cellular automata, this property translates to the fact that the global rule $F$ must be bijective to ensure that each global state of the cellular array has exactly one predecessor, and the inverse global mapping must also be a CA, that is, $F^{-1}$ has to be defined by a local rule. If these two requirements are fulfilled, then the corresponding infinite CA is called \emph{reversible}.

Hedlund~\cite{hedlund69} and Richardson~\cite{richardson72} independently proved that an infinite CA is reversible if and only if its global rule is bijective. In other words, bijectivity in a CA is sufficient to grant the property that the inverse global rule $F^{-1}$ is both shift-invariant and continuous. However, this result does not give a constructive proof to find the inverse global rule $F$. Indeed, even characterizing the diameter of the inverse local rule in a reversible CA is still an open problem, as shown by Czeizler and Kari~\cite{czeizler05}.

The relationship between bijectivity and reversibility is less straightforward in the case of finite CA. If we start from a local rule $f$ that generates a reversible infinite CA, then we can conclude that the same rule will give rise to a reversible PBCA for any length $n \in \N$ of the cellular array. This is because the set of spatially periodic configurations is a proper subset of the full-shift space $A^{\Z}$, and it is exactly the subset where PBCA act upon.
Conversely, if we know that a local rule $f$ induces a bijective global rule on a PBCA of a certain length $n \in \N$, then the inverse global rule is not necessarily defined by a local rule, nor is it the case that the global rule stays bijective for different lengths of the PBCA under the same local rule.

Local rules that generate bijective global rules only for certain lengths $n \in \N$ of the CA array and whose inverses cannot be described by local rules are also called \emph{globally invertible}. An example is the $\chi$ transformation used in the {\sc Keccak} sponge construction for hash functions~\cite{keccak}, which corresponds to a CA of length $n=5$ and is defined by the local rule of diameter $d=3$ with the Wolfram code 210. The offset of this CA is $\omega=0$, meaning that each cell applies rule 210 over itself and the two cells to its right to update its state. The algebraic expression of $\chi$ is:
\begin{equation}
	\label{eq:chi}
	\chi(x_1,x_2,x_3) = x_1 \oplus (x_2(1 \oplus x_3)) \enspace .
\end{equation}
In other words, the cell in position 1 flips its state if and only if the logical AND of $x_2$ and the complement of $x_3$ is true. Daemen~\cite{daemen95} showed that the rule 210 is globally invertible since it induces a bijective global rule only for odd lengths of the cellular array.
In particular, the inverse mapping can be specified by a sequential algorithm that takes as input a vector of odd length and a ``seed'' value, which is basically a single component of the preimage. Then, the other components of the preimage are determined by ``leaps'' of length two by going leftwards with respect to the seed. Since the vector has an odd length and periodic boundary conditions, each of the remaining components can be determined using a single seed. 
The fact that a preimage seed can always be found for any configuration of odd length shows why the resulting CA is indeed reversible. More details about the inversion procedure with seeds and leaps for rule $\chi$ can be found in~\cite{daemen95}.

On the other hand, a local rule that induces a bijective global function for all finite lengths $n \in \N$ of the cellular array is called \emph{locally invertible}. Using a topological argument that relies on the compactness of the full shift space~\cite{kari18}, it can be shown that locally invertible rules induce bijective global functions also on infinite CA. Hence, from the discussion above, it follows that locally invertible rules are exactly those defining reversible CA, where the inverse global rule $F^{-1}$ is determined by a local rule for all lengths $n \in \N$ of the cellular array. In what follows, we consider searching for locally invertible rules as an optimization problem, focusing on marker CA.

We conclude this section by mentioning that de Bruijn graphs can be used to study the reversibility of CA. Sutner~\cite{sutner91} devised an algorithm to decide whether any CA rule induces a reversible infinite CA $F$ by using its de Bruijn graph representation $G_{DB}(F)$. The algorithm starts by first computing the Cartesian product $G^2_{DB}(F)$ of the graph by itself; then, the CA is reversible if and only if the only non-trivial strongly connected components of the product $G^2_{DB}(F)$ coincides with its diagonal.

\subsection{Marker CA}
\label{subsec:mark-ca}
Even though reversibility in one-dimensional CA can be decided by utilizing Sutner's algorithm mentioned in the previous section, up to now, only a few classes of reversible CA are known in the literature (see, e.g.,~\cite{kari18}). These classes are usually defined in terms of particular properties of the local rule so that a subset of the rules satisfying them can generate a reversible CA. In this section, we describe the class of \emph{marker CA} that are the focus of the main contributions of this paper in later sections.

A \emph{marker CA} (also known as a \emph{complementing landscape CA}~\cite{toffoli90}) can be defined as a CA having a local rule that always flips the bit of the cell in position $\omega$ (i.e., the one whose state is being updated) whenever the cells in its neighborhood take on a particular pattern, or \emph{marker}. Otherwise, the cell stays in its current state. The set of patterns defining a local rule of a marker CA can be formalized through the concept of a \emph{landscape}:
\begin{definition}
	\label{def:landsc}
	Let $d, \omega \in \N$ with $\omega < d$. A \emph{landscape} of width $d$ and center $\omega$ is a string $L = l_0 l_1 \cdots l_{\omega-1} \star l_{\omega+1} \cdots l_{d-1}$ where $l_i \in \{0,1,-\}$ for all $i \neq \omega$. 
\end{definition}

The $\star$ symbol in a landscape $L$ indicates the \emph{origin} of the neighborhood in the local rule (i.e., the cell whose state is being updated), and consequently, occurring at position $\omega$. The $-$ symbol represents a ``don't care'', meaning that the corresponding cell can be either in the state $0$ or $1$. Thus, landscapes can be considered as a restricted form of regular expressions over the binary alphabet $\{0,1\}$, where the ``don't care'' symbol stands for the regular expression $(0+1)$ (i.e., both $0$ and $1$ match).

A local rule of a marker CA is described by one or more landscapes, all having the same width $d$ and center $\omega$. In the multiple landscape case, a cell is flipped if its neighborhood partakes on any of the patterns included in the union $\bigcup_{i=1}^k L_i$ of the landscapes $L_1,\cdots, L_k$ defining the local rule. As an example, observe that the transformation $\chi$ used in Keccak, whose definition is recalled in Eq.~\eqref{eq:chi}, is a marker rule. Indeed, it can be seen that the cell $x_1$ flips its state if and only if $x_2$ and $x_3$ are equal to $1$ and $0$, respectively. Therefore, rule $\chi$ is defined by the single landscape $\star10$.

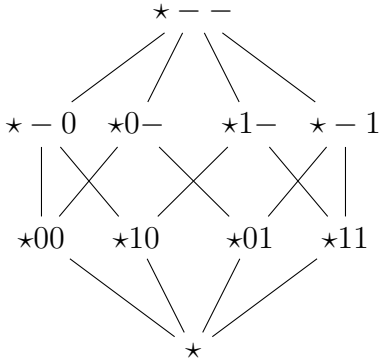
\begin{figure}[t]
	\centering
	\large
	\begin{tikzpicture}
	\node (max) at (0,4) {$\star--$};
	
	\node (a) at (-2,2.5) {$\star-0$};
	\node (a1) at (-0.75,2.5) {$\star0-$};
	\node (a2) at (0.75,2.5) {$\star1-$};
	\node (a3) at (2,2.5) {$\star-1$};
	
	\node (b) at (-2,1) {$\star00$};
	\node (b1) at (-0.75,1) {$\star10$};
	\node (b2) at (0.75,1) {$\star01$};
	\node (b3) at (2,1) {$\star11$};
	
	\node (min) at (0,-0.5) {$\star$};
	
	\draw (min) -- (b);
	\draw (min) -- (b1);
	\draw (min) -- (b2);
	\draw (min) -- (b3);
	
	\draw (a) -- (b);
	\draw (a) -- (b1);
	
	\draw (a1) -- (b);
	\draw (a1) -- (b2);
	
	\draw (a2) -- (b1);
	\draw (a2) -- (b3);
	
	\draw (a3) -- (b2);
	\draw (a3) -- (b3);
	
	\draw (max) -- (a);
	\draw (max) -- (a1);
	\draw (max) -- (a2);
	\draw (max) -- (a3);
	
	\end{tikzpicture}
	\label{fig:poset}
	\caption{Hasse diagram for the compatibility poset (partially ordered set) with $d=3$ and $\omega=0$. The landscape $\star10$ defines the rule $\chi$ introduced in~\cite{daemen95}.}
\end{figure}

It is possible to define a \emph{partial order} $\le_C$ over the set of landscapes. Namely, given two landscapes $L=l_0 \cdots l_{d-1}$ and $M = m_0 \cdots m_{d-1}$ with the same width $d$ and center $\omega$, we define
\begin{equation}
	\label{eq:part-ord}
	L\le_C M \Leftrightarrow l_i = m_i \textrm{ or } l_i \in \{0,1\} \textrm{ and } m_i = -
\end{equation}
for all $0\le i \le d-1$. 
Intuitively, this partial order describes the ``generality'' of a landscape: the more ``don't care'' symbols it has, the more patterns it contains. The bottom of this partial order is the trivial landscape $\star$, which corresponds to the identity rule (i.e., each cell copies its state without looking at its neighbors). Above this minimal element are the \emph{atomic landscapes}, which do not contain any ``don't care'' symbols, describing only single patterns. Finally, the top element is the landscape composed only of ``don't care'' symbols, which includes all possible patterns; the corresponding rule coincides with the complement of the identity, that is, the rule where each cell flips its state no matter what pattern its neighbors partake on. In what follows, we refer to $\le_C$ as the \emph{compatibility} partial order relation. In particular, we call two landscapes $L_1, L_2$ with the same width $d$ and center $\omega$ \emph{compatible} if $L_1 \le_C L_2$ or $L_2 \le_C L_1$. Otherwise, if $L_1$ and $L_2$ are not comparable with respect to~$\le_C$, we say that they are \emph{incompatible}. As an example, Figure~\ref{fig:poset} reports the diagram of the compatibility relation for $d=3$ and $\omega=0$.

The compatibility order relation can be used to characterize a subset of reversible marker CA, namely those of the \emph{conserved landscape} type. In such CA, a cell that is in a particular landscape $L$ defined by the local rule will still be in the \emph{same} landscape upon application of the global rule. This property can be formalized by requiring that the cells in the neighborhood are in landscapes that are incompatible with $L$, as shown in the following result proved in~\cite{toffoli90}:
\begin{lemma}
	\label{lm:loc-inv}
	Let $f: \F_2^d \rightarrow \F_2$ be a local rule of a marker CA defined by a set of $k$ landscapes $L_1,\cdots,L_k$ of width $d$ and center $\omega$. Further, for all $i \in \{1,\cdots,k\}$ let $M_{i,0}, \cdots, M_{i,\omega-1}, M_{i,\omega+1},\cdots,M_{i,d-1}$ be the set of $d-1$ landscapes associated to the neighborhood of $L_i$. Then, if $M_{i,j}$ is incompatible with all landscapes $L_1,\cdots,L_k$ for all $i \in \{1,\cdots,k\}$ and $j \in \{0,\cdots,\omega-1,\omega+1,\cdots,d-1\}$, rule $f$ induces a locally invertible marker CA.
\end{lemma}

When the conditions of Lemma~\ref{lm:loc-inv} are fulfilled, $f$ is named a \emph{conserved landscape} rule. Toffoli and Margolus noted that a conserved landscape local rule induces an \emph{involution}, i.e., the global rule of the resulting marker CA is its own inverse~\cite{toffoli90}. This is because any cell being in one of the marker landscapes will still be in the same landscape after applying the local rule. Therefore, after a further application of the local rule, the cell will go back to its initial state.

Conserved landscape rules define a particular type of reversible CA since all cycles have a length of $2$. Daemen argued that such CA could be useful in those cryptographic applications where both the encryption and decryption functions are implemented in hardware~\cite{daemen95}. It is also possible to relax the conditions of Lemma~\ref{lm:loc-inv} by allowing the landscapes of the local rule to partially \emph{overlap} one another~\cite{toffoli90}. In this case, a cell in a landscape defined by the local rule will be in any other landscape defined by the local rule after applying the global rule. As a consequence, the resulting marker CA can exhibit more complex behaviors with longer cycle lengths.

To better illustrate the idea, we provide an example of the only single conserved landscape rule of diameter $d=4$ (up to complement and reflection of the input), originally discovered by Patt~\cite{patt72}:
\begin{example}
	\label{ex:cons-land}
	Let $d=4$ and $\omega=1$, and let $f:\F_2^4 \rightarrow \F_2$ be the local rule defined by the single landscape $L = 0\star 10$. The tabulation depicted in Figure~\ref{fig:tab-cl} shows that all three landscapes of the neighboring cells are incompatible with $L$. In particular, when $x_i$ is in landscape $L$, then:
	\begin{compactenum}
		\item Cell $x_{i-1}$ is in landscape $- \star -1$, which is incompatible with $0 \star 10$ as there is a mismatch in position $3$.
		\item Cell $x_{i+1}$ is in landscape $- \star 0-$, which is incompatible with $0 \star 10$ as there is a mismatch in position $2$.
		\item Cell $x_{i+2}$ is in landscape $1 \star - -$, which is incompatible with $0 \star 10$ as there is a mismatch in position $0$.
	\end{compactenum}
	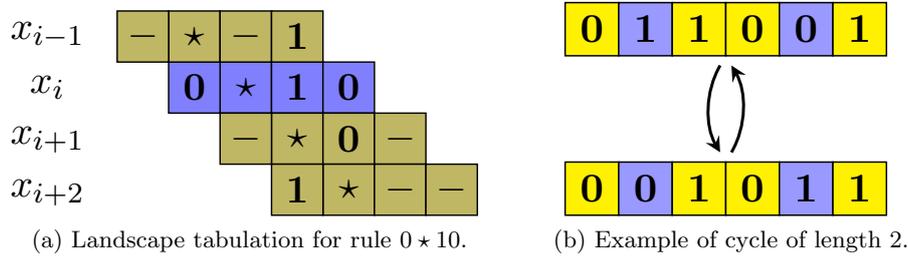
\begin{figure}[t]
		\centering
		\subfloat[Landscape tabulation for rule $0 \star 10$.\label{fig:tab-cl}]{
			\centering
			\resizebox{.5\textwidth}{!}{%
				\begin{tikzpicture}
				[->,auto,node distance=1.5cm, empt node/.style={font=\sffamily,inner
					sep=0pt,minimum size=0.3cm}, rect
				node/.style={rectangle,draw,fill=olive!60,font=\bfseries,minimum size=0.5cm, inner
					sep=0pt, outer sep=0pt}, blue
				node/.style={rectangle,draw,fill=blue!50,font=\bfseries,minimum size=0.5cm, inner
					sep=0pt, outer sep=0pt}]
				
				\node [blue node] (c1) {$\star$};
				\node [blue node] (c0) [left=0cm of c1] {0};
				\node [blue node] (c2) [right=0cm of c1] {1};
				\node [blue node] (c3) [right=0cm of c2] {0};
				\node [rect node] (d1) [above=0cm of c0] {$\star$};
				\node [rect node] (d0) [left=0cm of d1] {$-$};
				\node [rect node] (d2) [right=0cm of d1] {$-$};
				\node [rect node] (d3) [right=0cm of d2] {1};
				\node [rect node] (a1) [below=0cm of c2] {$\star$};
				\node [rect node] (a0) [left=0cm of a1] {$-$};
				\node [rect node] (a2) [right=0cm of a1] {0};
				\node [rect node] (a3) [right=0cm of a2] {$-$};
				\node [rect node] (b1) [below=0cm of a2] {$\star$};
				\node [rect node] (b0) [left=0cm of b1] {1};
				\node [rect node] (b2) [right=0cm of b1] {$-$};
				\node [rect node] (b3) [right=0cm of b2] {$-$};
				
				\node [empt node] (e1) [left=1cm of c0] {$x_i$};
				\node [empt node] (e0) [above=0.2cm of e1] {$x_{i-1}$};
				\node [empt node] (e2) [below=0.2cm of e1] {$x_{i+1}$};
				\node [empt node] (e3) [below=0.2cm of e2] {$x_{i+2}$};
				\end{tikzpicture}
			}
		}%
		\phantom{MM}%
		\subfloat[Example of cycle of length $2$.\label{fig:ex-rev-orb}]{
			\centering
			\resizebox{.35\textwidth}{!}{%
				\begin{tikzpicture}
				[->,auto,node distance=1.5cm, empt node/.style={font=\sffamily,inner
					sep=0pt,minimum size=0.1cm}, rect0
				node/.style={rectangle,draw,fill=orange,font=\bfseries,minimum size=0.5cm, inner
					sep=0pt, outer sep=0pt},
				rect1 node/.style={rectangle,draw,fill=yellow,font=\bfseries,minimum size=0.5cm, inner
					sep=0pt, outer sep=0pt},
				blue node/.style={rectangle,draw,fill=blue!40,font=\bfseries,minimum size=0.5cm, inner
					sep=0pt, outer sep=0pt}]
				
				\node [rect1 node] (c1) {0};
				\node [blue node] (c2) [right=0cm of c1] {1};
				\node [rect1 node] (c3) [right=0cm of c2] {1};
				\node [rect1 node] (c4) [right=0cm of c3] {0};
				\node [blue node] (c5) [right=0cm of c4] {0};
				\node [rect1 node] (c6) [right=0cm of c5] {1};
				
				\node [rect1 node] (a1) [below=1cm of c1] {0};
				\node [blue node] (a2) [right=0cm of a1] {0};
				\node [rect1 node] (a3) [right=0cm of a2] {1};
				\node [rect1 node] (a4) [right=0cm of a3] {0};
				\node [blue node] (a5) [right=0cm of a4] {1};
				\node [rect1 node] (a6) [right=0cm of a5] {1};
				
				\path[thick,shorten >=3pt,shorten <=3pt,>=stealth]
				(c3.south east) edge[bend right] (a3.north east)
				(a3.north east) edge[bend right] (c3.south east);
				\end{tikzpicture}
			}
		}%
		\caption{A locally invertible CA defined by the single landscape $0\star 10$. Figure~\ref{fig:ex-rev-orb} displays an example of a cycle starting from the initial state $011001$. The two cells in blue are in the landscape $0\star 10$.}
	\end{figure}
\end{example}

\section{Related Works}
\label{sec:rel-works}

As far as we are aware, our work is the first one exploring the application of evolutionary algorithms to evolve reversible CA. Therefore, in this section we briefly discuss related works related to the use of EA to evolve shift-invariant transformations and related objects for other tasks, such as random number generation. For a somewhat outdated but very detailed overview of works using GA to evolve CA, we refer interested readers to~\cite{mitchell96}.

B\"{a}ck and Breukelaar used genetic algorithms to evolve behavior in CA and explored different neighborhood shapes~\cite{back05}. The authors showed that their approach works for different topologies and neighborhood shapes.
Sipper and Tomassini~\cite{sipper96} proposed a cellular programming algorithm to co-evolve the rule map of non-uniform CA for designing random number generators. 
With their approach, the authors managed to evolve good generators that exhibit behaviors similar to those from the previously described CAs. Additionally, the authors reported advantages stemming from a ``tunable'' algorithm for obtaining random number generators.

Picek et al. demonstrated that GP could be used to evolve CA rules suitable to produce S-boxes (nonlinear elements used in block ciphers) with good cryptographic properties~\cite{picek17}. This approach allowed finding optimal S-boxes for several sizes of practical importance. Interestingly, this is the first time that EA managed to obtain optimal S-boxes for larger sizes.
Next, Picek et al. used genetic programming to demonstrate that the S-boxes obtained from the CA rules could have good implementation properties~\cite{picek17a}. The authors concentrated on two S-box sizes, $4\times 4$ and $5\times 5$, and managed to find S-boxes with good latency, area, and power consumption.
Subsequently, Mariot et al. conducted a more detailed analysis of the S-boxes based on CA, and they proved the best possible values for relevant cryptographic properties when CA rules of a certain size are used~\cite{mariot19}. The authors also used GP to experimentally validate their findings and reverse engineer a CA rule from a given S-box.

Mariot et al. used EA to construct orthogonal Latin squares built from CA~\cite{mariot17}. 
The authors reported that GP could always generate orthogonal Latin squares, where the optimal solutions were mostly linear. On the other hand, when using GA, the results were significantly worse than GP in evolving orthogonal Latin squares, but the corresponding Boolean functions were always nonlinear.
Finally, Mariot et al. investigated the possibility of evolving Reversible Cellular Automata (RCA). The authors considered three optimization strategies and obtained good results~\cite{10.1007/978-3-030-44094-7_8}. 
%This research is an extension of that work but contains new and significantly different experiments as detailed in Section~\ref{sec:intro}. %More precisely, we 1) consider additional evolutionary algorithms, 2) investigate larger problem instances, and 3) evaluate cellular automata with a fixed size of the offset $\omega$.

We note that the evolution of CA rules for cryptographic purposes is connected with the evolution of Boolean functions with good cryptographic properties. This direction is rather well-explored, and there are multiple works considering various evolutionary approaches, see, e.g.,~\cite{picek17b,mariot19a,JAKOBOVIC2021107327}.

\section{Optimizing the Reversibility of CA}
\label{sec:opt-rev}
In this section, we propose two different approaches to formulate the search of reversible CA as an optimization problem that can be tackled with evolutionary algorithms.
The first approach considers the case of generic reversible CA and exploits the de Bruijn graph representation.
However, we argue that such an approach is not suitable to find CA rules that are locally invertible since it relies on a specific length of the cellular array.
We thus introduce a second approach, where we consider only the class of conserved landscape rules that do not have this problem.
For this reason, we focus on this particular approach for our experimental evaluation in the next sections.

\subsection{Generic Reversible CA}
\label{subsec:gen-ca}
As mentioned in Section~\ref{subsec:rev-ca}, Sutner's algorithm shows that the reversibility property in CA is decidable in the one-dimensional case~\cite{sutner91}. Although any decision problem can be easily turned into an optimization one (e.g., by defining a cost function that evaluates to 1 if a solution is optimal and 0 otherwise), this usually results in fitness landscapes that are too hard to be explored by any optimization algorithm. For this reason, we investigate an optimization approach that departs from Sutner's product graph construction, even though it is still based on the de Bruijn graph representation. In particular, the approach is loosely inspired by a technique used by the authors in~\cite{mariot19}, where bijective $n\times n$ S-boxes defined by CA of diameter $d=n$ are reverse-engineered to verify whether they can be expressed by local rules of smaller diameter.

Instead of searching for a local rule and then optimizing its reversibility, this optimization approach takes the opposite direction: we start from a bijective mapping $F: \F_2^n \to \F_2^n$ and then tweak it so that it corresponds to the global rule of a finite reversible CA. In this way, the genotype representation is quite simple since it suffices to define a candidate solution as a permutation on a set of $2^n$ elements. Evolutionary algorithms such as permutation-based GA~\cite{larranaga99} or modified versions of GP (like the one proposed in~\cite{picek15}) can then be adopted to variate a population of individuals such that the permutation constraint is preserved. Thus, the optimization objective becomes to find a permutation that, when interpreted as a bijective vectorial Boolean function $F: \F_2^n \to \F_2^n$, is defined by the application of a single local rule of diameter $d \le n$. However, this approach raises the question of defining a proper fitness function to evaluate candidate permutations.

This is where the de Bruijn graph representation comes into play: as noted in Section~\ref{subsec:bas-def}, a CA local rule corresponds to labeling on the edges of a de Bruijn graph. Hence, the idea is to start from a blank de Bruijn graph (i.e., without labels on its edges) and then fill it up by traversing each input $c \in \F_2^n$ of the permutation and then labeling the corresponding edges for each cell in the corresponding output $F(c)$. Here, \emph{traversing} an input configuration $c \in \F_2^n$ means to sweep a window of width $d$ on it (similarly to a convolution operator), using periodic boundary conditions. For each position $i$ of the window in $c \in \F_2^n$, we will observe a particular neighborhood configuration $x \in \F_2^d$, which coincides with an edge on the de Bruijn graph. The value of the cell in position $i$ of $F(c)$ will be the output $f(x)$ of the local rule reputed to define the permutation $F$, or equivalently the label of the edge corresponding to $x$. Clearly, by sweeping each input of a random permutation, it is  likely that a neighborhood configuration $x \in \F_2^d$ will get distinct outputs, or  equivalently that the labeling on the corresponding edge is \emph{inconsistent}. Therefore, the permutation $F$ is defined by a reversible CA if and only if the labeling is \emph{consistent} after traversing all inputs in $\F_2^n$, i.e., if and only if each edge has a unique label. On the contrary, if some edge $(v_1,v_2) \in E$ gets more than one label after traversing all inputs, it means that the input neighborhood $v_1 \odot v_2 \in \F_2^d$ does not have a single output, or equivalently that the truth table does not correspond to a Boolean function.

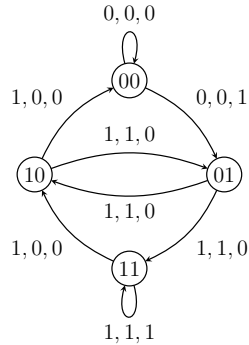
\begin{figure}[t]
	\centering
	\subfloat[Permutation truth table.\label{fig:tt-perm}]{
		\centering
		\raisebox{1.45cm}{
			\begin{tabular}{ccc|ccl}
				\hline
				$x_1$ & $x_2$ & $x_3$ & \multicolumn{3}{c}{$F(x_1,x_2,x_3)$} \\
				\hline
				$0$   & $0$  & $0$   &  \ $0$   & $0$ & $0$    \\
				$0$   & $0$  & $1$   &  \ $0$   & $1$ & $0$    \\
				$0$   & $1$  & $0$   &  \ $0$   & $1$  & $1$ \\
				$0$   & $1$  & $1$   &  \ $1$   & $0$  & $1$ \\
				$1$   & $0$  & $0$   &  \ $0$   & $0$  & $1$ \\
				$1$   & $0$  & $1$   &  \ $1$   & $1$  & $0$ \\
				$1$   & $1$  & $0$   &  \ $1$   & $0$  & $0$ \\
				$1$   & $1$  & $1$   &  \ $1$   & $1$  & $1$ \\
				\hline
			\end{tabular}
		}
	}%
	\phantom{MMMM}
	\subfloat[De Bruijn graph labeling.\label{fig:db-perm}]{
		\centering
		\Large
		\resizebox{3.5cm}{!}{
			\begin{tikzpicture}
			[->,auto,node distance=0.5cm, every loop/.style={min distance=10mm},
			empt node/.style={font=\sffamily,inner sep=0pt,outer sep=0pt},
			circ node/.style={circle,thick,draw,font=\sffamily\bfseries,minimum
				width=0.8cm, inner sep=0pt, outer sep=0pt}]
			
			% Nodes
			\node [empt node] (e1) {};
			\node [circ node] (n00) [above=1.75cm of e1] {$00$};
			\node [circ node] (n01) [right=1.75cm of e1] {$01$};
			\node [circ node] (n10) [left=1.75cm of e1] {$10$};
			\node [circ node] (n11) [below=1.75cm of e1] {$11$};       
			
			% Edges
			\draw [->, thick, shorten >=0pt,shorten <=0pt,>=stealth] (n00) 
			edge[bend left=20] node (f5) [above right]{$0,0,1$} (n01);
			\draw [->, thick, shorten >=0pt,shorten <=0pt,>=stealth] (n01)
			edge[bend left=20] node (f5) [below right]{$1,1,0$} (n11);
			\draw [->, thick, shorten >=0pt,shorten <=0pt,>=stealth] (n11)
			edge[bend left=20] node (f5) [below left]{$1,0,0$} (n10);
			\draw [->, thick, shorten >=0pt,shorten <=0pt,>=stealth] (n10)
			edge[bend left=20] node (f5) [above left]{$1,0,0$} (n00);
			\draw[->, thick, shorten >=0pt,shorten <=0pt,>=stealth] (n10) edge[bend
			left=20] node (f1) [above]{$1,1,0$} (n01);
			\draw[->, thick, shorten >=0pt,shorten <=0pt,>=stealth] (n01)
			edge[bend left=20] node (f2) [below]{$1,1,0$} (n10);
			\draw[->, thick, shorten >=0pt,shorten <=0pt,>=stealth] (n00) edge[loop
			above] node (f3) [above]{$0,0,0$} ();
			\draw[->, thick, shorten >=0pt,shorten <=0pt,>=stealth] (n11) edge[loop
			below] node (f4) [below]{$1,1,1$} ();
			\end{tikzpicture}
		}
	}
	\caption{Example of inconsistent labeling from a permutation $F:\F_2^3 \to \F_2^3$.}
	\label{fig:ex-lab}
\end{figure}

We illustrate the above reasoning with a small example. Let $n=d=3$ and $\omega=0$. Therefore, we are interested in finding a permutation $F: \F_2^3 \to \F_2^3$ such that $F$ is the global rule of a CA defined by a local update rule $f: \F_2^3 \to \F_2$ of diameter $3$. Figure~\ref{fig:ex-lab} reports the truth table of a random permutation $F$ and the corresponding labeling on the de Bruijn graph.
As it can be seen, each edge has three associated labels. This is because we are considering the case of a finite cellular array of length $n=d$ with periodic boundary conditions. Each neighborhood configuration is seen exactly $n$ times while traversing the whole truth table of $F$. For instance, the loops on $00$ and $11$ are seen three times, respectively, by considering all cyclic rotations of the input vectors $000$ and $111$. Incidentally, these two labelings are the only consistent ones in the whole graph since we always observe $0$ for $000$ and $1$ for $111$. Thus, by narrowing our attention only on these two input configurations, the permutation could in principle be defined by a CA, since each cell always gets either a $0$ or a $1$ when its neighbors are respectively set either all to $0$ or to $1$.

However, for all remaining edges, we can see that the labelings are always inconsistent since both $0$ and $1$ occur. For example, the edge $(00, 01)$ is labeled with two $0$s and a single $1$. This means that in two cases, the neighborhood $001 = 00 \odot 01$ is mapped to $0$ for the next state of a cell, while it is mapped to $1$ in a third case. Thus, the permutation cannot be defined by a CA local rule.

The above observations can be generalized to formulate a fitness function that assigns fitness $0$ to every candidate permutation inducing a consistent labeling or equivalently that defines a reversible CA. For instance, one possibility could be to define the fitness of a permutation as the sum of the Hamming distances of the labelings on the edges from the vector $(0,0,\cdots, 0)$ or $(1,1,\cdots,1)$, and then minimize such a sum.
However, following this approach to evolve a population of reversible CA with evolutionary algorithms is not devoid of problems. Indeed, we can remark two main issues arising from the adoption of such a fitness function:
\begin{compactitem}
	\item The most obvious drawback is that, as discussed in Section~\ref{subsec:rev-ca}, finding a local rule that induces a permutation for a fixed length $n$ of the cellular array does not imply that the same rule will give a reversible CA for other lengths. Indeed, such a rule will likely be globally but not locally invertible. Globally invertible rules are also interesting, as exemplified by the $\chi$ transformation used in Keccak. Still, using this optimization approach to find globally invertible rules would not give a straightforward way to characterize the array lengths for which the corresponding CA is reversible.
	\item The fitness function depends both on the length $n$ of the cellular array and the diameter $d$ of the sought local rule. In the above example, we made the simplifying assumption that $n=d$. However, in principle, one could consider if a particular permutation of length $n$ is defined by a CA local rule for any diameter $d\le n$. This would significantly increase the number of possible fitness functions to consider (i.e., one for each possible diameter).
\end{compactitem}
Consequently, the approach based on evolving permutations and optimizing their consistency on the edge labeling of the de Bruijn graph is not very scalable and practical to find local rules that generate reversible CA for any length of the array. Since we are mostly interested in locally invertible rules, we will not investigate further this optimization approach in later sections.

\subsection{Conserved Landscape CA}
\label{subsec:cons-ca}

We now focus on a different optimization perspective by considering only the class of conserved landscape reversible CA. Lemma~\ref{lm:loc-inv} tells us that to find a marker CA that is locally invertible, we need to define a set of landscapes $L_1,\cdots,L_k$, in such a way that their associated neighborhood landscapes are incompatible with them. This suggests the following idea to turn the search of conserved landscape rules into an optimization problem: given the landscape specification of a local rule, \emph{count} the number of compatible landscape pairs, and minimize it. Using the partial order relationship that we defined in
Section~\ref{subsec:mark-ca}, this is equivalent to minimize the number of comparable pairs of landscapes. An optimal solution is a set of landscapes that are all mutually incompatible (including the neighborhood landscapes), or equivalently an \emph{antichain} of elements in the poset induced by $\le_C$. These observations lead to the following optimization problem:
\begin{problem}
	\label{prob:stat}
	Let $d,\omega \in \N$ with $0 < \omega < d-1$. Find a set of landscapes $L_1,\cdots,L_k$ of width $d$ and center $\omega$, such that for all $i \in \{1,\cdots,k\}$ and for all $j \in \{0,\cdots,d-1\}$, the neighborhood landscape $M_{i,j}$ associated to $L_i$ is incompatible with all other landscapes $L_1,\cdots,L_k$, that is $M_{i,j} \not\le_C L_{t}$ and $L_{t} \not\le_C M_{i,j}$ for all $t \in \{1,\cdots,k\}$.
\end{problem}
In the rest of this section, we will first address how to obtain the landscape representation of a local rule from its truth table, and then we will define the fitness function to be minimized for Problem~\ref{prob:stat}.

\subsubsection{Genotype Representation for Marker CA}
\label{subsec:enc}

The first question arising from Problem~\ref{prob:stat} is how to represent local rules of marker CA so that they can be evolved by the variation operators of GA and GP. In particular, GA usually works on a bitstring encoding of the candidate solutions of an optimization problem, while GP relies on a tree representation. Hence, directly using the landscape specification of a marker CA rule does not seem a natural choice for encoding the genotype.

Recall from Eq.~\eqref{eq:chi} that the $\chi$ rule was defined as the XOR of the leftmost cell (which also coincides with the cell being updated) with the AND between the second cell and the complement of the third cell. This observation can be generalized to any local rule of marker CA as follows. Let $L_1,\cdots, L_k$ be a set of landscapes of diameter $d$ and center $\omega$ defining a local rule $f:\F_2^d \rightarrow \F_2$. Additionally, let
$\mathcal{L} = \bigcup_{i=1}^k L_i$ be the union of the landscapes. Then, a cell $x_i$ in a marker CA equipped with rule $f$ will flip its state if and only if the neighborhood $x_{i-\omega} \cdots x_{i-1}\!\star\!x_{i+1}\cdots x_{i+d-1-\omega}$ belongs to $\mathcal{L}$. Excluding the origin $\star$ of the neighborhood, we obtain a vector of $d-1$ variables that describes the states of the cells surrounding $x_i$. Consider now all $2^{d-1}$ possible assignments to this vector, and let $g: \F_2^{d-1} \rightarrow \F_2$ be the Boolean function defined as:
\begin{equation}
	\label{eq:gen-func}
	g(x_{i-\omega} \cdots x_{i-1}x_{i+1}\cdots x_{i+d-1-\omega}) =
	\begin{cases}
		1, \ \textrm{if } x_{i-\omega} \cdots
		x_{i-1}\!\star\!x_{i+1}\cdots x_{i+d-1-\omega} \in \mathcal{L} \enspace \\
		0, \ \textrm{otherwise } \enspace ,
	\end{cases}
\end{equation}
for all $x_{i-\omega} \cdots x_{i-1}x_{i+1}\cdots x_{i+d-1-\omega} \in \F_2^{d-1}$. 
In other words, function $g$ outputs $1$ if and only if the configuration featured by the cells surrounding $x_i$ belongs to the union of landscapes $\mathcal{L}$, when the origin $\star$ is inserted at position $\omega$. Then, it follows that the local rule $f$ can be expressed as
\begin{equation}
	\label{eq:mark-ca-gen}
	f(x_{i-\omega} \cdots x_{i-1}x_ix_{i+1}\cdots x_{i+d-1-\omega}) = x_i \oplus
	g(x_{i-\omega} \cdots x_{i-1}x_{i+1}\cdots x_{i+d-1-\omega}) \enspace ,
\end{equation}
for all configurations of $x_{i-\omega} \cdots x_{i-1}x_ix_{i+1}\cdots x_{i+d-1-\omega} \in
\F_2^d$. Hence, the algebraic form of the local rule of a marker CA can be expressed as the XOR of the cell in the origin with the \emph{generating  function} $g$ computed on the surrounding cells. Indeed, $g$ evaluates to $1$ if and only if the neighborhood takes on any of the landscapes in $\mathcal{L}$, and in this case $x_i$ will flip its state.

Consequently, we can reduce the representation of the local rule $f$ of a marker CA to its generating function $g$, since we can compute $f$ by simply XORing the output of $g$ with the value of $x_i$. Since $g$ can be any Boolean function of $d-1$ variables, it follows that we can represent the genotype of a candidate solution to our optimization problem with the commonly used Boolean genotype encodings for GA
and GP.
In particular:
\begin{compactitem}
	\item For GA, the genotype of a candidate solution is a bitstring of length $2^{d-1}$, representing the output of the truth table of $g$.
	\item For GP, the genotype is a tree where the terminal nodes represent the input variables of $g$ (i.e., the state of the cells surrounding the origin of the neighborhood), while the internal nodes are Boolean operators combining the values received from their child nodes and propagating their output to their parent node. The output of the root node will be the output of the whole generating function $g$.
\end{compactitem}

\subsubsection{Fitness Functions}
\label{subsec:fit}
At the beginning of this section, we informally introduced the idea to steer the search of conserved landscape rules by counting the number of compatible pairs of landscapes. However, given that the genotype handled by GA and GP is an encoding of the generating function $g$, we first need to translate this representation to the landscape specification.

Suppose that we have the truth table of the generating function $g$. In the GA case, this corresponds exactly to the genotype of an individual. For GP, we can easily recover it by evaluating the Boolean tree of an individual over all possible input vectors $x \in \F_2^{d-1}$. Let $supp(g) = \{x \in \F_2^{d-1}: g(x) \neq 0 \}$ be the \emph{support} of $g$,
i.e., the set of input vectors over which $g$ evaluates to $1$. By construction, the elements of $supp(g)$ coincide with all the patterns that the cells surrounding the origin must feature to flip the state of the central cell. Thus, to obtain the list of atomic landscapes, it  suffices to insert the origin symbol $\star$ in position $\omega$ to each vector of the support. Of course, some of these patterns could be described in a more ``compact'' way with more general landscapes that also use the ``don't care'' symbol. For example, if $supp(g) = \{101, 111\}$ and the center is $\omega=1$, then the two atomic landscapes $1\!\star\!01$ and $1\!\star\!11$ can be described by the single landscape $1\!\star\!-1$, where we substituted the central variable with a ``don't care'' symbol\footnote{This method can be generalized using the following greedy procedure. Let $supp(g)$ be the support of the generating function, and remove $x, y \in supp(g)$ such that their  Hamming distance is $1$. Then, insert in $supp(g)$ the landscape $L$ that has the same symbols as $x$ and $y$, except for the single position in which they differ, where $L$ has a ``don't care'' symbol $-$. Repeat this procedure until no further replacements can be performed (i.e., all pairs of landscapes in $supp(g)$ have the Hamming distance higher than $1$).}.

However, the set of atomic landscapes obtained from the support suffices to check if a rule is of the conserved landscape type or not. It is not difficult to see that two landscapes containing ``don't care'' symbols are incompatible if and only if all the atomic landscapes that they describe are incompatible between themselves. This means that we can directly use the support of the generating function to \emph{count} the number of pairs of compatible landscapes. Given that we want to minimize such a number in order to get a conserved landscape rule, we define the following objective function:
\begin{definition}
	\label{def:fitness}
	Let $g: \F_2^{d-1}\rightarrow \F_2$ be a generating function of a marker CA rule $f: \F_2^d\rightarrow \F_2$ of diameter $d$ and offset $\omega$, and let $supp(g)$ be its support. Further, let $L_1,\cdots,L_k$ be the set of atomic landscapes obtained by adding the origin symbol $\star$ in position $\omega$ to each vector in $supp(g)$, and for each $i \in \{1,\cdots,k\}$ let $M_{i,0}, \cdots, M_{i,\omega-1}, M_{i,\omega+1},\cdots,M_{i,d-1}$ be the set of neighborhood landscapes associated to $L_i$, obtained through the tabulation procedure. Then, the fitness function value of $g$ is defined as follows:
	\begin{equation}
		\label{eq:obj1}
		obj_1(g) = \sum_{i \in [k], j \in [d-1]_\omega} \sum_{t \in [k]} comp(M_{i,j},L_t) \enspace ,
	\end{equation}
	where $[k] = \{1,\cdots,k\}$,
	$[d-1]_\omega = \{0,\cdots,\omega-1,\omega+1,\cdots,d-1\}$, and the function $comp(\cdot,\cdot)$ returns $1$ if the two landscapes passed as arguments are compatible, and $0$ otherwise.
\end{definition}
Hence, the objective function loops over all neighborhood landscapes $M_{i,j}$ induced by each atomic landscape $L_i$, compares each of these neighborhood landscapes with all atomic landscapes $L_1,\cdots, L_k$ through the function $comp(\cdot,\cdot)$, and adds $1$ whenever a compatible pair is found. Therefore, the function $obj_1$ measures the degree of compatibility of a set of atomic landscapes induced by the support of a generating function $g$. The optimization objective is thus to \emph{minimize} $obj_1$, with $obj_1(g) = 0$ corresponding to an optimal solution where all neighborhood landscapes are incompatible with the atomic landscapes, and thus the latter define a conserved landscape rule.

Secondly, a good indicator of the complexity of the dynamical behavior of a marker CA is the \emph{Hamming weight} of its generating function $g$, i.e., the cardinality of its support. This metric can also be used as a proxy for the utility of a marker CA in cryptography since it is related to the \emph{nonlinearity} of the resulting vectorial Boolean function~\cite{carlet21}. Given a generating function $g$, we thus define a second optimization objective function as follows:
\begin{equation}
	\label{eq:obj2}
	obj_2(g) = | supp(g) | \enspace .
\end{equation}

In the optimization of these objectives, we experimented using three optimization scenarios.
The first one, which we denote as the \textit{single-objective} scenario, included only the minimization of the reversibility objective. The fitness function for the first scenario is then simply defined as:
\begin{equation}
	\label{eq:fitness}
	fit_1(g) = obj_1(g) \enspace ,
\end{equation}
where the optimization goal is minimization.

As it became apparent quite early in our experiments that this goal is very easily attainable with both representations, we modified the fitness function so the evolution could generate more distinct solutions with different Hamming weights. 
This is made possible simply by maximizing the Hamming weight value, but only for solutions that already obtained a conserved landscape solution, i.e., those for which the first objective is already minimized. At the same time, whenever an algorithm reaches a solution with a higher Hamming weight, every such individual is added to a set of distinct solutions reported at the end of each run.

Therefore, in the second scenario, which is denoted as \textit{lexicographic optimization}, we are interested in maximizing the Hamming weight while retaining an optimal value of $obj_1$. For this reason, we define a second fitness function for this particular case as follows:
\begin{equation}
	\label{eq:fit2}
	fit_2(g) = 
	\begin{cases}
		obj_1 \ , & \textrm{if } obj_1 > 0 \enspace , \\
		-obj_2 \ , & \textrm{if } obj_1 = 0 \enspace . \\
	\end{cases}
\end{equation}
Stated otherwise, with the second fitness function, we still minimize $obj_1$ until we reach a reversible rule, and after that, we minimize the opposite of the Hamming weight (thus, equivalently, we are maximizing $obj_2$).

Finally, we included a multi-objective approach to investigate the interaction between the reversibility of a marker CA rule and the Hamming weight of its generating function. In the \textit{multi-objective} scenario, we \textit{minimized} the reversibility objective $obj_1$ and \textit{maximized} the Hamming weight as defined by $obj_2$ in Equation~\eqref{eq:obj2}.

\section{Experimental Evaluation}
\label{sec:exp}
In this section, we present the experimental setting and results obtained by applying GA and GP on Problem~\ref{prob:stat}. We start by performing an exhaustive exploration of all conserved landscape rules up to diameter $d=6$, which is still computationally feasible. Next, we use the findings obtained from the exhaustive search to formulate our research questions and lay down our experimental settings. Finally, we present the results of our parameter tuning and discuss them in light of our research questions.

\subsection{Preliminary Exhaustive Search}
\label{subsec:exh-search}
As noted in Section~\ref{subsec:enc}, the local rule of a marker CA of diameter $d$ can be identified with its generating function $g$ of $d-1$ variables, computed on the neighborhood cells surrounding the origin since the state of the central cell is XORed with the result of $g$. Given a diameter $d \in \N$, this means that we can define the phenotype space as the set $\mathcal{P}(d) = \{g: \F_2^{d-1} \rightarrow \F_2\}$ of all Boolean functions of $d-1$ variables. The genotype space, on the other hand, will correspond to the set of all binary strings of length $2^{d-1}$ specifying the truth tables $\Omega_g$ of the generating functions in $\mathcal{P}(d)$. For GP, it will be the space of all Boolean trees whose terminals represent the $d-1$ input variables and the internal nodes represent Boolean operators.

Since the number of Boolean functions of $d-1$ variables is $2^{2^{d-1}}$, the phenotype space $\mathcal{P}(d)$ can be exhaustively searched for reversible marker CA rules up to diameter $d=6$, since there are at most $2^{32}\approx 4.3 \cdot 10^{9}$ generating functions to check for the conserved landscape property. As far as we know, an exhaustive search of reversible marker CA rules has been carried out only by Patt~\cite{patt72}, who considered diameters up to $d=4$. For completeness, Table~\ref{tab:exhaustive} reports the numbers of conserved-landscape rules we found by exhaustively searching the sets of generating functions up to $d=6$ for each possible value of $\omega$, along with the length of the truth table ($2^{d-1}$), the size of the phenotype space (\#$\mathcal{P}(d)$), and the observed Hamming weights. Recall that the Hamming weight of the generating function corresponds to the number of atomic landscapes over which a cell flips its state. We excluded from the count the identity rule, which copies the state of the central cell since it is trivially reversible for any diameter. Further, we halved the numbers of the remaining rules since, if a rule is of the conserved landscape type, then its complement is too. As a general remark, one can see from Table~\ref{tab:exhaustive} that the number of conserved landscape rules is much smaller than the size of the whole generating function set for any offset $\omega$. Another interesting observation is that the highest numbers of conserved landscape rules are always found when $\omega$ corresponds to the center of the neighborhood or to its immediate left or right (if $d$ is even). Indeed, the extreme cases are $\omega=0$ and $\omega=d$, where no conserved landscape rules exist. As noted in~\cite{daemen95}, if the offset is on either the leftmost or rightmost cell of the neighborhood, then any landscape is always compatible with at least another one. Also, the fact that the distributions of conserved landscape rules are symmetrical to the center of the neighborhood is backed by the results proved in~\cite{toffoli90}, where reversible marker rules in different offsets are shown to be symmetric under rotations and reflection. Finally, the number of the observed Hamming weights is quite limited since, for the largest considered instance of diameter $d=6$, we only found reversible rules defined by at most three landscapes, which are thus not very useful for cryptographic and reversible computing purposes.
\begin{center}
	\begin{table}
		\caption{Numbers of conserved landscape rules found by exhaustive search, up to equivalence by complement and excluding the trivial identity rule. }
		\centering
		\begin{tabular}{ccp{1cm}ccp{0.8cm}}
			\toprule
			$d$ & $2^{d-1}$ & \#$\mathcal{P}(d)$    & $\omega$ & \#REV & Weights \\ 
			%\midrule
			%\multirow{3}{*}{$3$} & \multirow{3}{*}{$4$}  & \multirow{3}{*}{$16$}             & $0$ & $0$  & $-$     \\
			%                     &                       &                                   & $1$ & $0$  & $-$     \\
			%                     &                       &                                   & $2$ & $0$  & $-$     \\
			\midrule
			\multirow{4}{*}{$4$} & \multirow{4}{*}{$8$}  & \multirow{4}{*}{$256$}            & $0$ & $0$  & $-$     \\
			&                       &                                   & $1$ & $1$  & $1$     \\
			&                       &                                   & $2$ & $1$  & $1$     \\
			&                       &                                   & $3$ & $0$  & $-$     \\
			\midrule
			\multirow{5}{*}{$5$} & \multirow{5}{*}{$16$} & \multirow{5}{*}{$65\,536$}        & $0$ & $0$  & $-$     \\
			&                       &                                   & $1$ & $2$  & $1$     \\
			&                       &                                   & $2$ & $5$  & $1,2$   \\
			&                       &                                   & $3$ & $2$  & $1$     \\
			&                       &                                   & $4$ & $0$  & $-$     \\
			\midrule
			\multirow{6}{*}{$6$} & \multirow{6}{*}{$32$} & \multirow{6}{*}{$4.3 \cdot 10^9$} & $0$ & $0$  & $-$     \\
			&                       &                                   & $1$ & $8$  & $1,2$   \\
			&                       &                                   & $2$ & $23$ & $1,2,3$ \\
			&                       &                                   & $3$ & $23$ & $1,2,3$ \\
			&                       &                                   & $4$ & $8$  & $1,2$   \\  
			&                       &                                   & $5$ & $0$  & $-$     \\
			\bottomrule
		\end{tabular}
		\label{tab:exhaustive}
	\end{table}
\end{center}

\subsection{Research Questions}
\label{subsec:rq}
The empirical observations obtained from the exhaustive search experiments presented in the previous section prompted us with three research questions:
\begin{compactitem}
	\item {\bfseries RQ1:} Does the limited number of conserved landscape rules with respect to the search space size imply a difficulty for evolutionary algorithms to find them?
	\item {\bfseries RQ2:} Do there exist conserved landscape rules of a larger diameter that are useful for cryptographic and reversible computing applications, i.e., having larger Hamming weights with respect to the size of the generating function truth table?
	\item {\bfseries RQ3:} Is there a trade-off between the reversibility of a marker CA rule (as measured by the objective function $obj_1$ defined in Section~\ref{subsec:fit}) and its Hamming weight (as defined by the second objective $obj_2$)?
\end{compactitem}
Although these research questions are inherited from the conference version of this work~\cite{10.1007/978-3-030-44094-7_8}, we emphasize that here they are explored from a different perspective, especially concerning the first two. In particular, in our previous conference paper, the offset $\omega$ was fixed to the neighborhood center, i.e., $\omega = \lfloor (d-1)/2\rfloor$. The reason for that choice was that, as shown in Table~\ref{tab:exhaustive}, most of the reversible rules are found when the offset is closer to the center.

On the other hand, in this work, we consider the situation where the offset is fixed to $\omega=3$ for the experiments described in the next sections. The reason is twofold: first, by keeping the offset to a fixed value, one could reasonably expect that the difficulty for evolutionary algorithms to converge to an optimal solution increases even more by considering larger diameters than by placing $\omega$ near to the center. Indeed, increasing the diameter while keeping $\omega$ fixed means that the origin of the landscape rules gets farther from the center. Consequently, as experimentally observed through an exhaustive search, the number of optimal solutions becomes smaller, and this, in turn, likely affects the answers to RQ1 and RQ2 as discussed in our conference paper~\cite{10.1007/978-3-030-44094-7_8}. Further, in principle, one may assume that the trade-off between the compatibility fitness and the Hamming weight could change by considering an offset far from the center of the neighborhood, thereby potentially affecting also the answer to RQ3. Finally, the second reason for choosing a fixed $\omega$ in our investigation is more of a practical nature: in this way, we can adopt a more uniform experimental setting, especially concerning the parameter tuning phase described in Section~\ref{subsec:par-tun}.

\subsection{Experimental Settings}
\label{subsec:rq-es}

We utilized a genetic algorithm with truth table encoding and genetic programming with a tree-based representation to investigate the stated research questions. Both representations use the same selection scheme, a steady-state elimination tournament; in each iteration, three individuals are randomly selected from the population. A new solution is generated by applying crossover to the best two individuals from the tournament. The new individual undergoes mutation, subject to a predefined individual mutation rate, which is an algorithm parameter. Finally, the new individual replaces the worst one from the tournament, and the process is repeated. Each iteration produces one new individual and performs a single fitness evaluation.
Apart from the described method, we also experimented with an evolutionary strategy-based scheme, in which a number of offspring is generated using mutation only; however, preliminary experiments showed that this selection method produced inferior results for both representations. In the multi-objective approach, we used the well-known NSGA-II algorithm~\cite{deb02}.

For the truth table binary representation (GA), we employed one-point, two-point, and uniform crossover operators selected at random at each iteration. The mutation operator was a single bit-flip on a randomly selected position.
In the tree-based representation (GP), five crossover operators are used at random: simple subtree crossover, uniform crossover, size fair, one-point, and context preserving crossover. Similarly to GA, a single mutation operator was used, the subtree mutation, with a fixed mutation probability of 0.5~\cite{poli2008field}.

The function set used in the tree-based encoding included the binary operators AND, OR, XOR, XNOR, AND with the second input complemented, and the unary operator NOT. Additionally, we included the ternary function IF, which returns the second argument if the first one is true and the third one otherwise. We performed a tuning phase to investigate which subset of these functions provides the best results.

For each considered optimization scenario (single-objective where $fit_1$ is minimized, multi-objective where $obj_1$ and $obj_2$ are respectively minimized and maximized, and lexicographic optimization where $fit_2$ is minimized) we performed our experiments on the spaces of marker CA rules with diameter $7 \le d \le 15$. Therefore, with respect to our previous results reported in~\cite{10.1007/978-3-030-44094-7_8}, we extended our investigation with two additional diameter values. Each experiment was repeated for $50$ independent runs to obtain statistically reliable results, and each run was given a budget of $500\,000$ evaluations, which is the same as adopted in~\cite{10.1007/978-3-030-44094-7_8}. Indeed, as it will be clear in the next sections, such a budget proved to be more than sufficient to investigate our research questions, and we deemed unnecessary a larger one.

\subsection{Parameter Tuning}
\label{subsec:par-tun}
To set up the different parameters of the evolutionary algorithms employed for our experiments, we performed a tuning phase on the instance of marker CA rules with diameter $d=10$ and $\omega=3$. Recall, in our previous experiments presented in~\cite{10.1007/978-3-030-44094-7_8}, we carried out this phase on $d=7$ and $\omega=3$. While we already elaborated in Section~\ref{subsec:rq} why we chose an asymmetric offset for all our experiments, we tuned our evolutionary algorithms on a larger problem instance mainly for robustness reasons. Indeed, $d=7$ is the smallest instance where it makes sense to tune an evolutionary algorithm for this problem, since, for smaller values, the search space is limited enough that the problem can be easily solved by exhaustive search, as discussed in Section~\ref{subsec:exh-search}. Moreover, in this case, $\omega=3$ corresponds to the center of the neighborhood. Hence, we selected $d=10$ as a sufficiently representative instance of our new experimental setting since the offset is far enough from the peak of the distribution of optimal solutions occurring in the center.

In the GA case, we tuned the population size $p$ and the mutation probability $\mu$, with the population size ranging among the values $\{100, 200, 500\}$, while the mutation probability was in the range $\{0.3, 0.4, 0.5, 0.6, 0.7, 0.8, 0.9\}$.

For GP, besides the population size in the same range as GA, we tuned the maximum depth of the trees, considering the values in $\{5,7,9,11\}$. The motivation is that in our experiments in~\cite{10.1007/978-3-030-44094-7_8} we always set the maximum depth equal to $d-1$, mainly following a heuristic adopted in previous works on the optimization of Boolean functions~\cite{picek17b,mariot19a,JAKOBOVIC2021107327}. However, one could argue that by using these methods for large diameters such as $d\ge 10$, one could end up with very large trees, eventually making the investigation of the evolved solutions for interpretability more difficult. Likewise, one could also argue that a larger maximum depth could be beneficial to converge more rapidly on an optimal solution. For this reason, we experimented both with smaller and greater maximum depth with respect to the initial value $d-1$.

Finally, the third parameter that we tuned for GP is the set of Boolean operators used in the internal nodes of the trees. In~\cite {10.1007/978-3-030-44094-7_8}, we used a function set composed of four binary operators (AND, OR, XOR, and XNOR), one unary operator (NOT), and one ternary operator (IF). Again, the motivation for this choice was the previous experience with optimization problems related to Boolean functions solved using GP~\cite{picek17b,mariot19a,JAKOBOVIC2021107327}. However, as correctly remarked by one of the reviewers of~\cite{10.1007/978-3-030-44094-7_8}, such a set could easily induce the GP trees to bloat since, for example, XNOR is equivalent to the composition of NOT and XOR. Although bloat was already controlled in our previous experiments by adopting the maximum depth parameter, we decided to investigate this question more thoroughly. To this end, we started with a minimal set of operators such that any Boolean function can be expressed by a combination of them, i.e., AND, OR, and NOT. Then, we added to this minimal set the combinations of XOR, XNOR, AND with the second input complemented, and IF, retaining only the combinations that significantly improved the results.

For both GA and GP tuning, each parameter combination was tuned with a fitness budget of $100\,000$ evaluations, repeated in $30$ independent runs for statistical significance purposes. After each run, the fitness value of the best individual was recorded, thus obtaining a sample of $30$ observations that approximated the distribution of the best fitness for a particular parameter combination. Moreover, to select the parameter combinations to be used in our subsequent experiments, we performed a two-stage statistical analysis with non-parametric tests. First, we used the \emph{Kruskal-Wallis} test~\cite{kruskal52} to compare a group of parameter combinations all at once, using a significance level of $\alpha=0.05$. If no significant differences were observed, then another criterion for selecting the parameter combination to be used among those in the group was adopted (i.e., highest median). On the other hand, if the distributions were detected to be significantly different, we employed the \emph{Mann-Whitney U} test~\cite{mann47} to perform pairwise comparisons and determine the best parameter combination. In particular, the null hypothesis for the test was that the random variable of the fitness represented by the first distribution was better than that of the second distribution. Here, the definition of ``better'' depends on the context: when we performed the tuning for the single-objective versions of our algorithms where only the reversibility fitness function is used, then better corresponds to lower fitness values. For the lexicographic optimization approach, on the other hand, the objective is to maximize the Hamming weight while retaining reversibility. Hence, in this case, better corresponds to the higher Hamming weight values. The significance level was again set to $\alpha=0.05$, applying \emph{Bonferroni correction}~\cite{dunn61} since we performed multiple comparisons.

\subsubsection{GA Tuning}
\label{subsec:ga-tun}
For the GA tuning, we performed a complete sweep across all $3\times 7 = 21$ parameter combinations for population size and mutation rate, considering both the single-objective case (SOGA), where only $fit_1$ is minimized, and the lexicographic optimization approach (LEXGA), where $fit_2$ is minimized. Concerning SOGA, \emph{no differences were detected} during the parameter sweep: indeed, for each considered parameter combination, the best solution always reached an optimal fitness in all $30$ experimental runs. For this reason, we focused only on the lexicographic optimization approach, adopting the same parameter combination selected for LEXGA also for SOGA.

\begin{figure}
	\centering
	\includegraphics[scale=0.6]{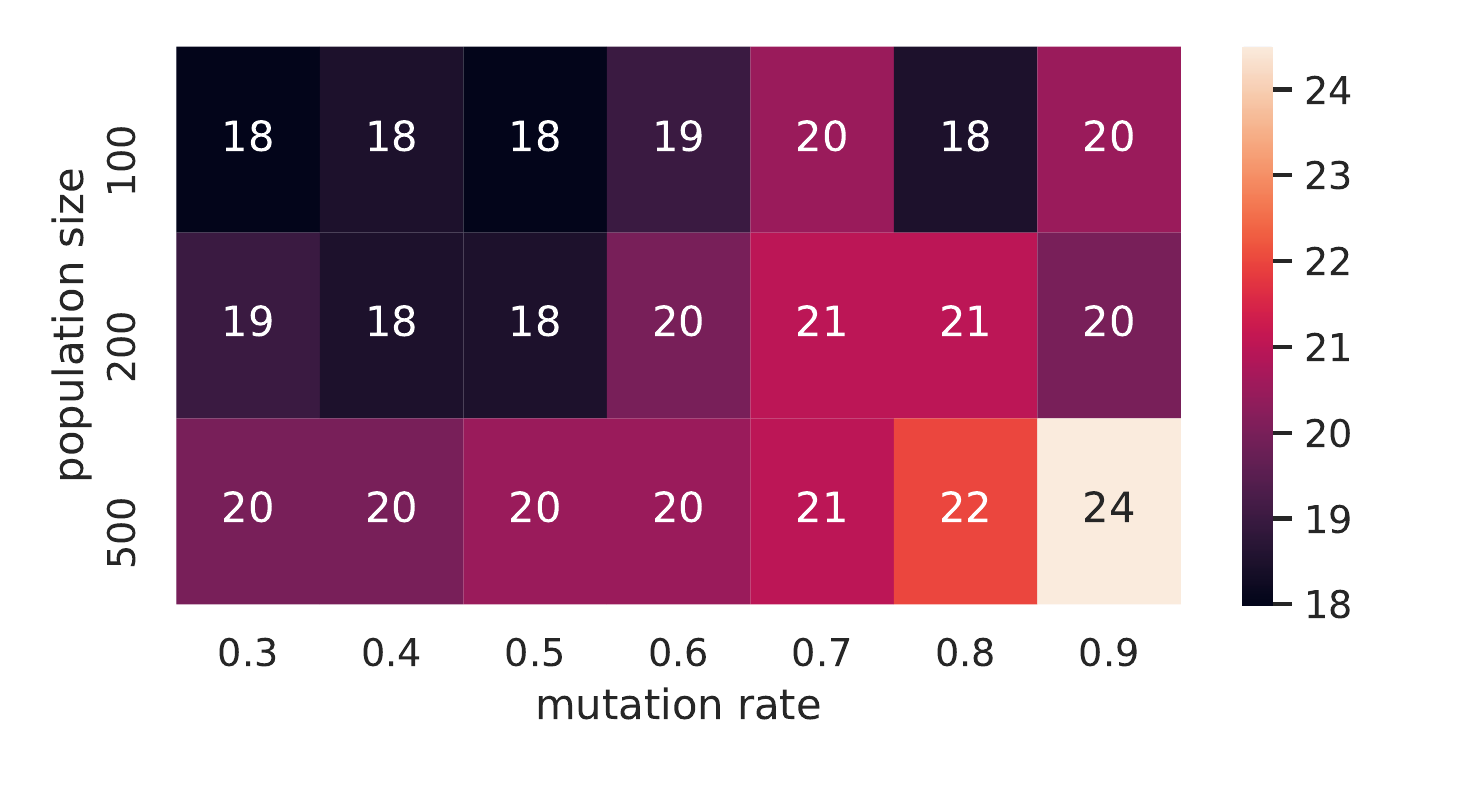}
	\caption{Heatmap for the tuning phase of LEXGA. The numbers inside the cells refer to the median fitness obtained by the best individual across all experimental runs.}
	\label{fig:gatuning}
\end{figure}

Figure~\ref{fig:gatuning} depicts the heatmap of the median best fitness obtained by LEXGA across all $21$ parameter combinations of population size and mutation probability. Note that we only show the Hamming weight being maximized as the second objective since the first objective was optimal in every case.
The color gradient already indicates an advantage in using large populations and high mutation rates. Indeed, after performing the Kruskal-Wallis test for all $21$ distributions, significant differences were detected. For this reason, we proceeded by performing pairwise comparisons through the Mann-Whitney U test. As a criterion to select the best parameter combination, we used a ranked tournament: each distribution was compared against all others, and if the Mann-Whitney U test rejected the null hypothesis (that is, the obtained $p$-value was below the corrected significance level), then a $+1$ was scored by the distribution, and the distribution scoring the highest number of points was then selected as a winner. This resulted in the combination $p=500$ and $\mu = 0.9$, since it achieved $20$ points (i.e., it had significant differences against all other combinations), while the second-best ones reached a consistently lower score of $7$. Incidentally, this analysis also confirmed the result suggested by the heatmap, although only the median best fitness was considered there. Therefore, both for LEXGA and SOGA, we selected a population size of $500$ individuals and a mutation probability of $0.9$.

\subsubsection{GP Tuning}
\label{subsec:gp-tun}
The number of parameter values to test for GP was $3$ for the population size, $4$ for the maximum depth, and $7$ for the subsets of operators. Checking all $84$ parameters combinations resulting from a grid search, as in the case of GA, would have implied a too large computational effort. Therefore, we decided to opt for a lexicographic tuning approach: first, we determined the best maximum depth among $\{5,7,9,11\}$ by keeping the population size fixed to $100$ individuals and using the minimal operators set of AND, OR, NOT. Then, we used the best maximum depth values and the same set of operators to tune the population size. Finally, we tuned the operators set by using the selected best population size and maximum depth.

Similar to the GA tuning, in the single-objective scenario (SOGP), no differences were observed since, in all the configurations, GP always obtained the optimal solution in every algorithm run. Therefore, we used the lexicographic scenario (LEXGP) to estimate the appropriate set of parameters.

Concerning the first phase (maximum depth tuning), significant differences were detected with the Kruskal-Wallis test on the set $\{5,7,9,11\}$. Using the Mann-Whitney U test with a ranked tournament as in the case of LEXGA, the best values for this parameter were 7 and 9, with no significant differences between them. For this reason, we kept them both for the next phase, where we analyzed all combinations of parameters for maximum depth in $\{7,9\}$ and population size in $\{100,200,500\}$. Once again, significant differences resulted from applying the Kruskal-Wallis tests on such distributions. Using the ranked tournament approach for the pairwise comparisons with the Mann-Whitney U test, we obtained $4$ remaining combinations, each achieving the same score. Among these $4$ remaining combinations, we selected the one with the highest median best fitness, i.e., $500$ individuals for the population size and maximum depth of $9$. Finally, for the last phase, where the tuning was performed by adding operators to the minimal set, no significant differences arose from the Kruskal-Wallis test. Hence, we again selected the combination with the highest median fitness. The final parameters combination selected for both LEXGP and SOGP was $p=500$, $d=9$ and operator set including AND, OR, NOT, and AND with second input complemented. In particular, since the selected maximum depth turned out to be equal to 9 while the tuning diameter was 10, we kept $d-1$ as a maximum depth for all other instances in the subsequent experiments, as done in our previous work~\cite{10.1007/978-3-030-44094-7_8}.

\section{Results}
\label{sec:results}

In this section, we present the results emerging from our experimental evaluation. First, we discuss the results for single-objective optimization, followed by those on multi-objective and lexicographic optimization. Finally, we analyze the diversity of the CA rules obtained in our experiments by using several metrics related to the number of unique solutions and the different Hamming weights.

\subsection{Single-objective Optimization Results}
\label{subsec:so}

\begin{figure}[t]
	\centering
	\includegraphics[width=0.8\textwidth]{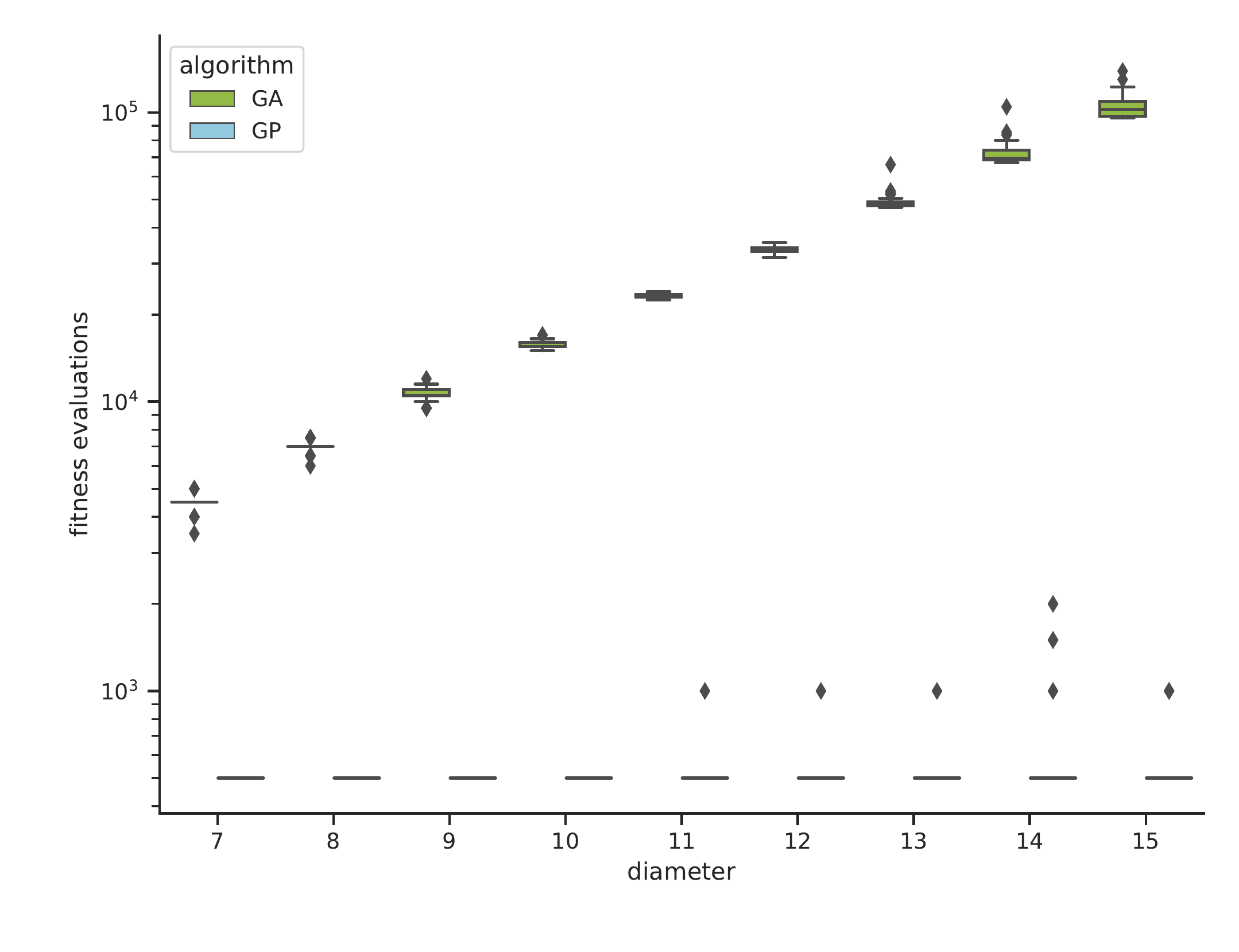}
	\caption{Comparison of fitness evaluations performed by SOGA and SOGP.}
	\label{fig:fiteval}
\end{figure}

Figure~\ref{fig:fiteval} gives results for the single-objective GA and GP considering the number of evaluations needed to reach the optimal value. For GP, we reach optimal fitness value for each dimension already in the initial population, making the results less interesting. A possible reason why GP shows such a behavior is that it is easier to guess a generating function that results in a reversible marker CA rule with a random algebraic expression than with a random string of bits, as in the case of GA.
Still, we require somewhat more evaluations for larger dimensions, indicating that it becomes slightly more difficult to guess optimal solutions randomly. 

For GA, we observe an exponential increase in the number of required fitness evaluations concerning the diameter sizes (remark that the fitness evaluations axis in the plot is in logarithmic scale). This indicates that larger problem instances are more difficult, but there should be no reason why GA would not work well on even larger diameters. A similar trend was also observed in our previous investigation~\cite{10.1007/978-3-030-44094-7_8}, although there, a much smaller population was used. Indeed, the number of fitness evaluations required with our current setting is consistently smaller than in our previous one, requiring less than 100\,000 fitness evaluations to converge for $d=13$. This seems to indicate that using a larger population is beneficial for GA.

\begin{figure}[h]
	\subfloat[d=12]{\includegraphics[scale=0.18]{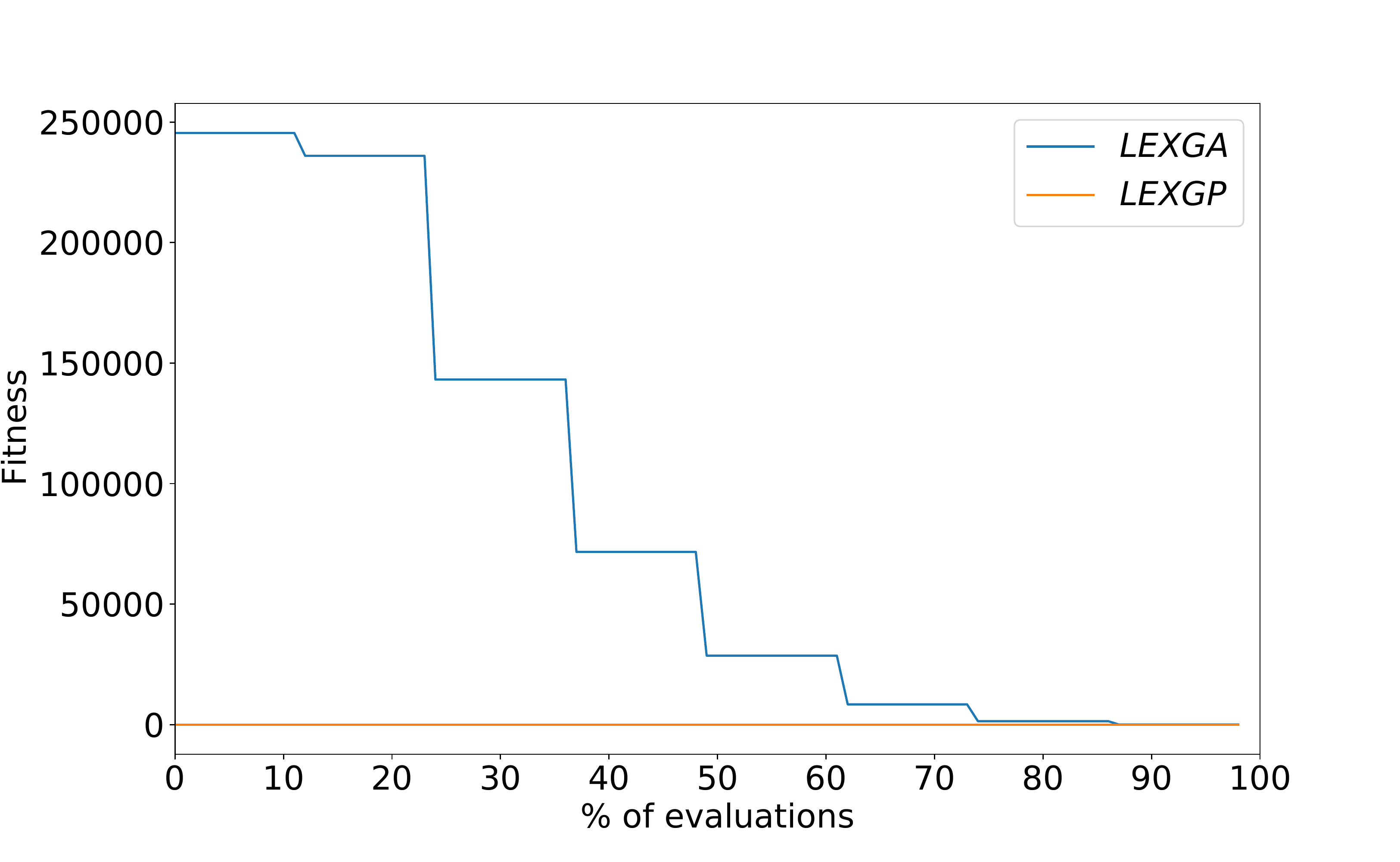} \label{fig:so_12}} 
	\subfloat[d=13]{\includegraphics[scale=0.18]{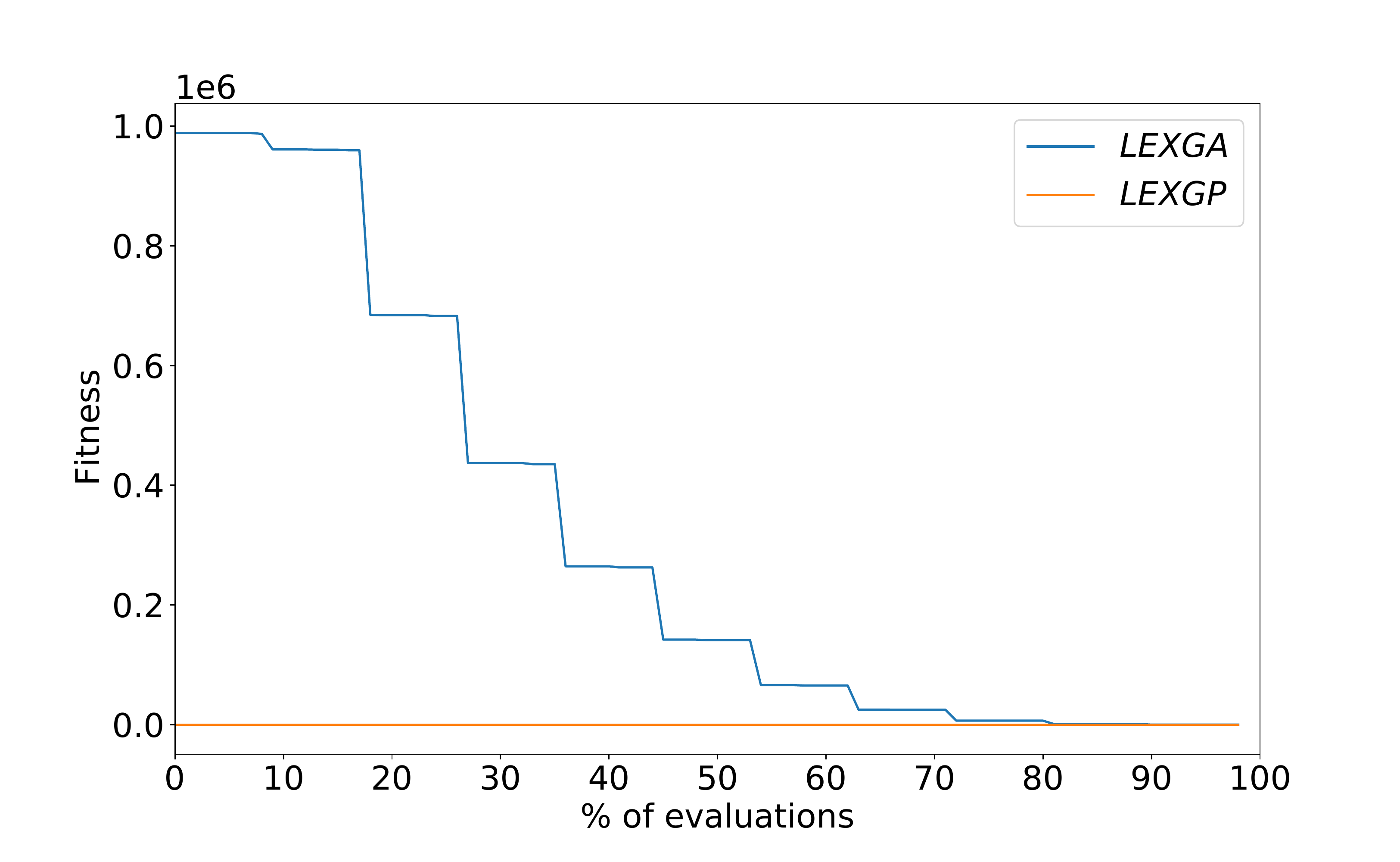} \label{fig:so_13}} \\
	\subfloat[d=14]{\includegraphics[scale=0.18]{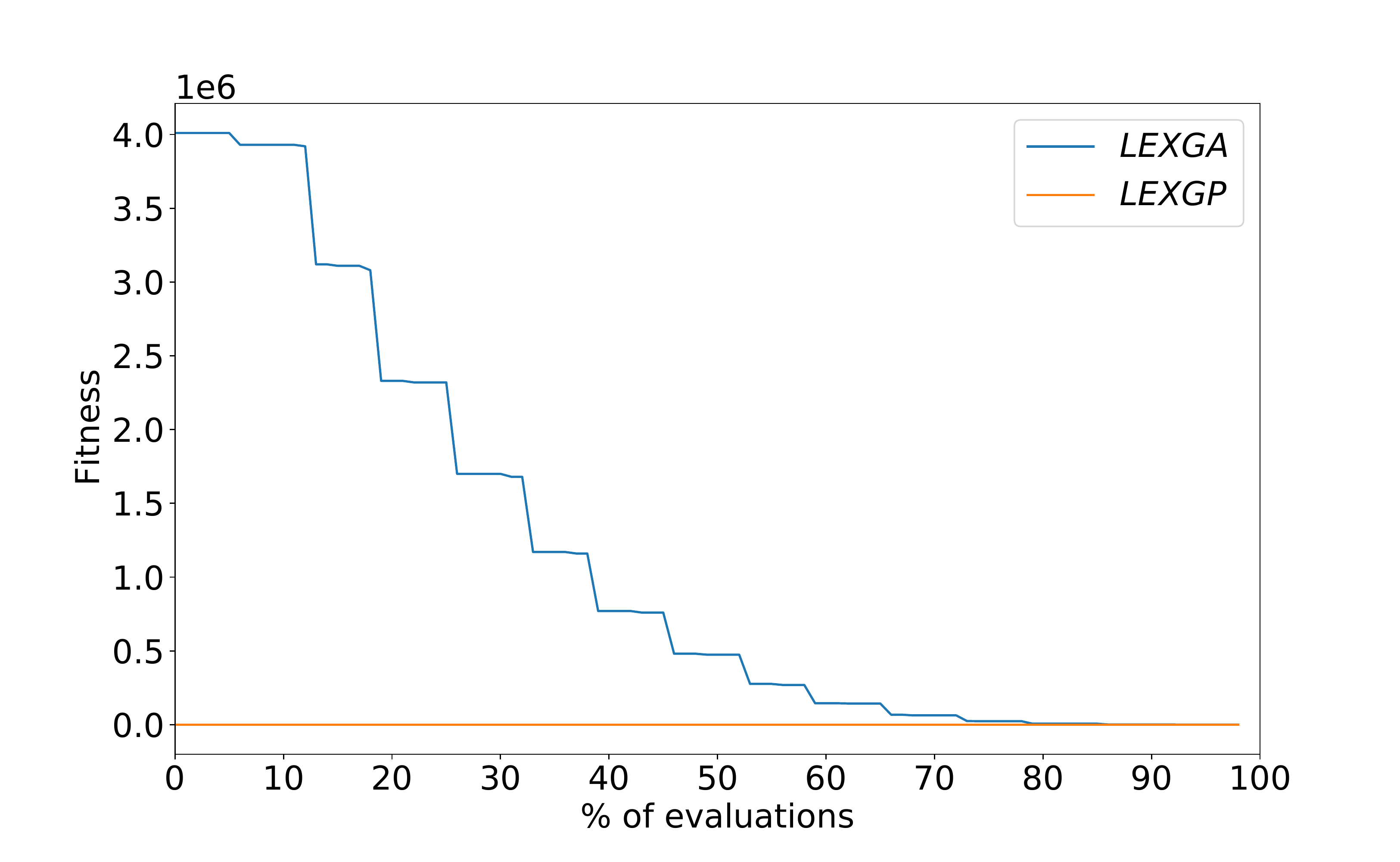} \label{fig:so_14}} 
	\subfloat[d=15]{\includegraphics[scale=0.18]{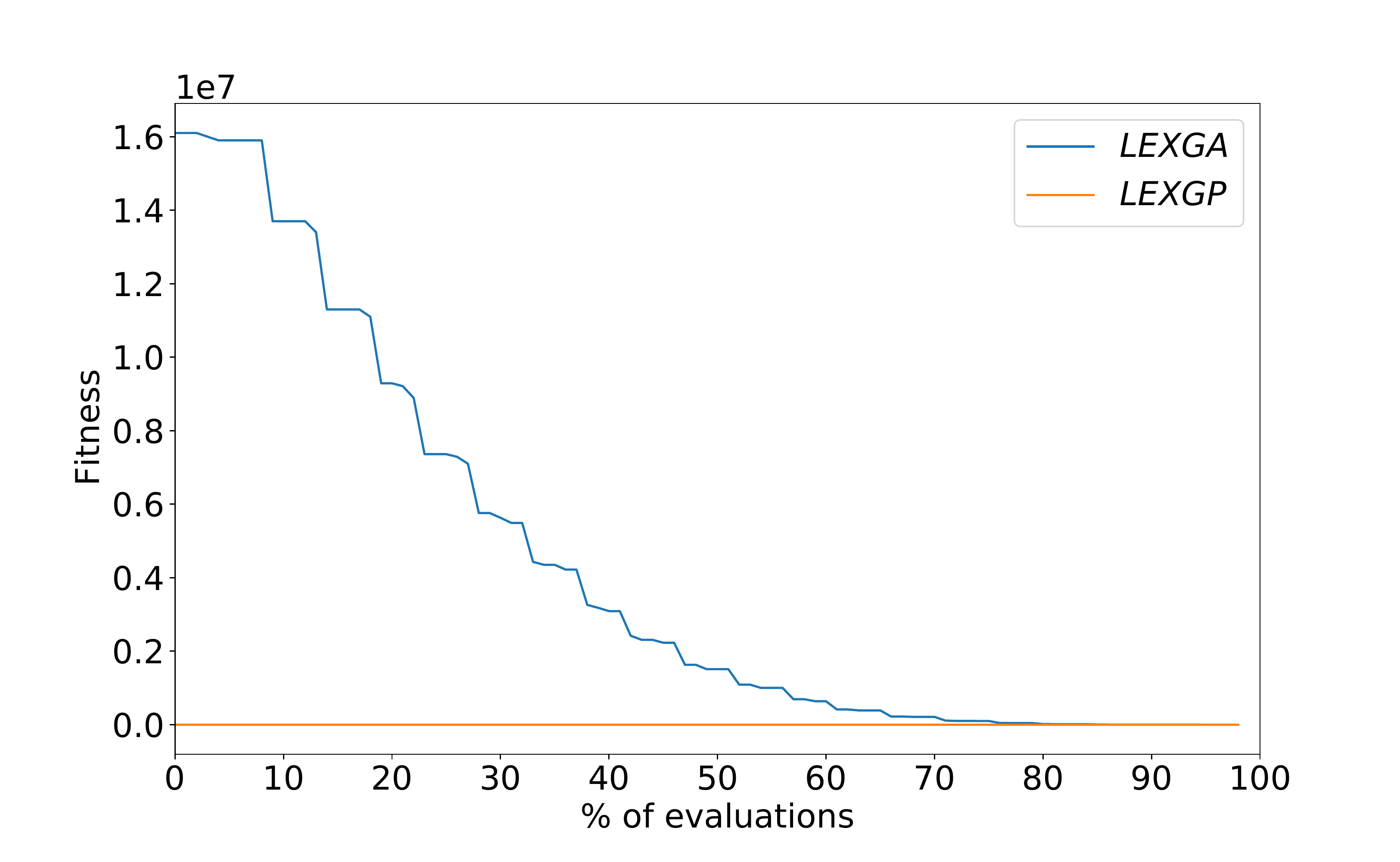} \label{fig:so_15}} 
	\caption{Single-objective convergence plots.}
	\label{fig:convergence_so}
\end{figure}

Next, in Figure~\ref{fig:convergence_so}, we display convergence plots for the GA and GP single-objective optimization algorithms. We plot the median best fitness results, focusing only on diameter size $d$ from 12 to 15, as smaller sizes show similar trends, but the optimization process becomes much easier. Notice that for GP, all cases show that the random initial population contains optimal solutions. On the other hand, GA starts with large fitness values but continuously improves them and reaches the optimal value after using around 70\% of the fitness evaluation budget allowed.

\subsection{Multi-objective Optimization Results}
\label{subsec:mo}
\begin{figure}[t]
	\centering
	\includegraphics[scale=0.6]{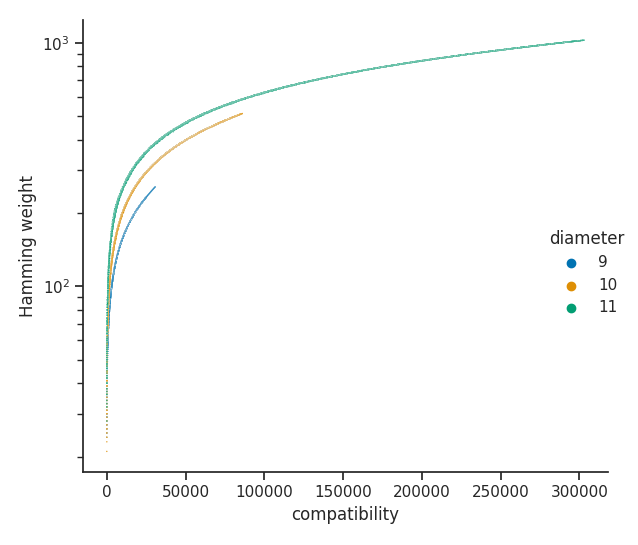}
	\caption{Pareto fronts for $9 \le d \le 11$ approximated by MOEA.}
	\label{fig:pareto}
\end{figure}

Figure~\ref{fig:pareto} depicts the Pareto fronts approximated by MOEA when minimizing the compatibility score (i.e., $obj_1$) and maximizing the Hamming weight (i.e., $obj_2$). For the sake of readability, we only report the fronts for $d=9,10,11$. The scale difference on the Hamming weight axis between one diameter size $d$ and the next one is so large that displaying all fronts between $d=7$ and $d=15$ would only make the larger ones visible, rendering indiscernible the smaller ones. However, this is not a serious issue since all fronts obtained in our experiments follow similar shapes.

The curves in Figure~\ref{fig:pareto} corroborates the previous findings reported in~\cite{10.1007/978-3-030-44094-7_8}: the closer a CA marker rule is to be of the conserved landscape type, the lower the Hamming weight of its generating function must be. The first extreme case occurs when the rule achieves an optimal compatibility score of 0 (i.e., the rule is reversible), with very small Hamming weights observed (see also Section~\ref{sec:disc} for an overview of the possible Hamming weights when adopting a lexicographic optimization approach). On the other side, one can see that the compatibility fitness reaches its highest values when the Hamming weight is maximal, and in particular, it is about half the length of the generating function truth table. Hence, this indicates that marker CA rules with \emph{balanced} generating functions (whose truth tables are composed of an equal number of 0s and 1s) are the farthest possible from being reversible under the conserved landscape definition.

\subsection{Lexicographic Optimization Results}
\label{subsec:lex}

\begin{figure}[t]
	\subfloat[d=12]{\includegraphics[scale=0.18]{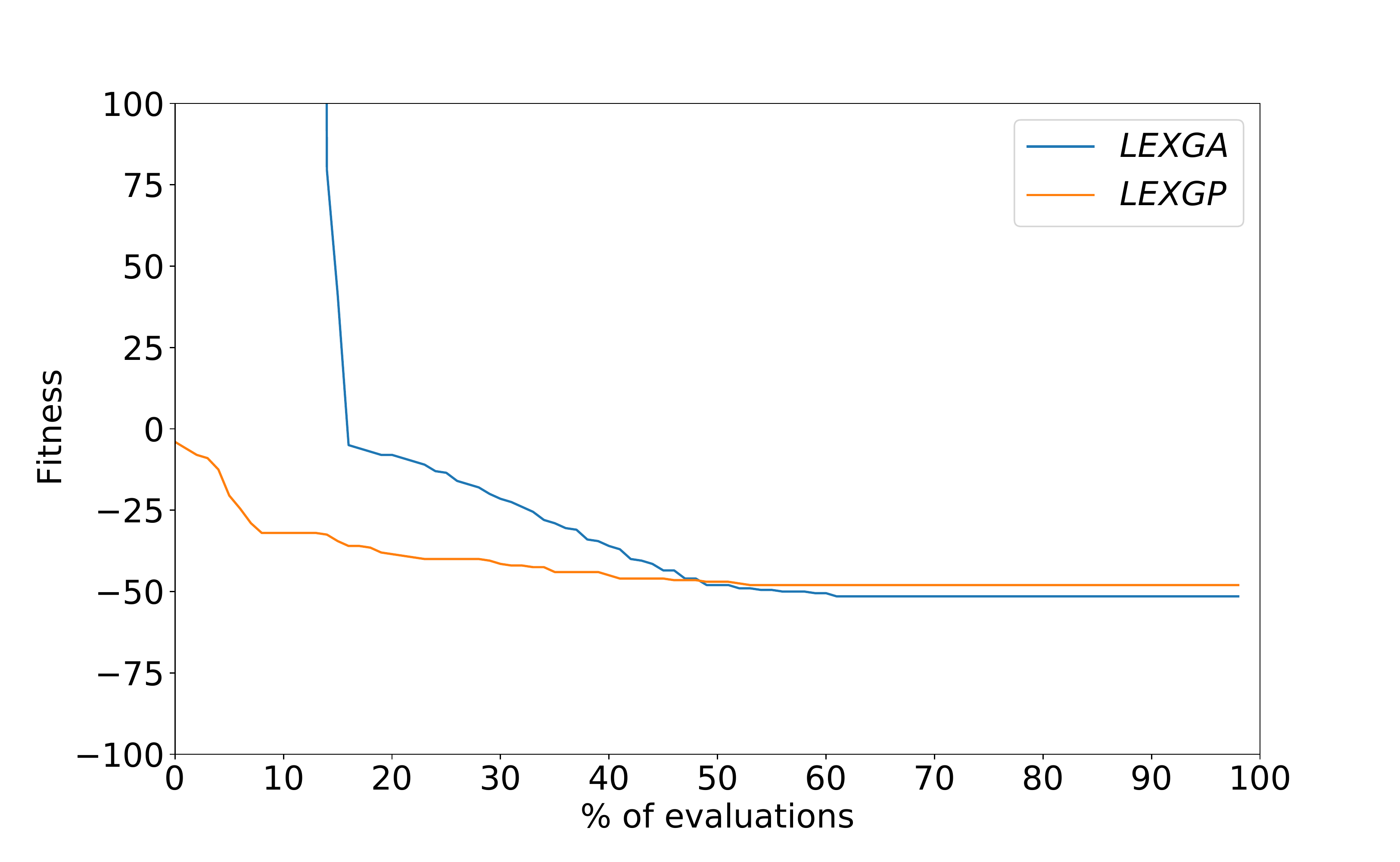} \label{fig:lex_12}} 
	\subfloat[d=13]{\includegraphics[scale=0.18]{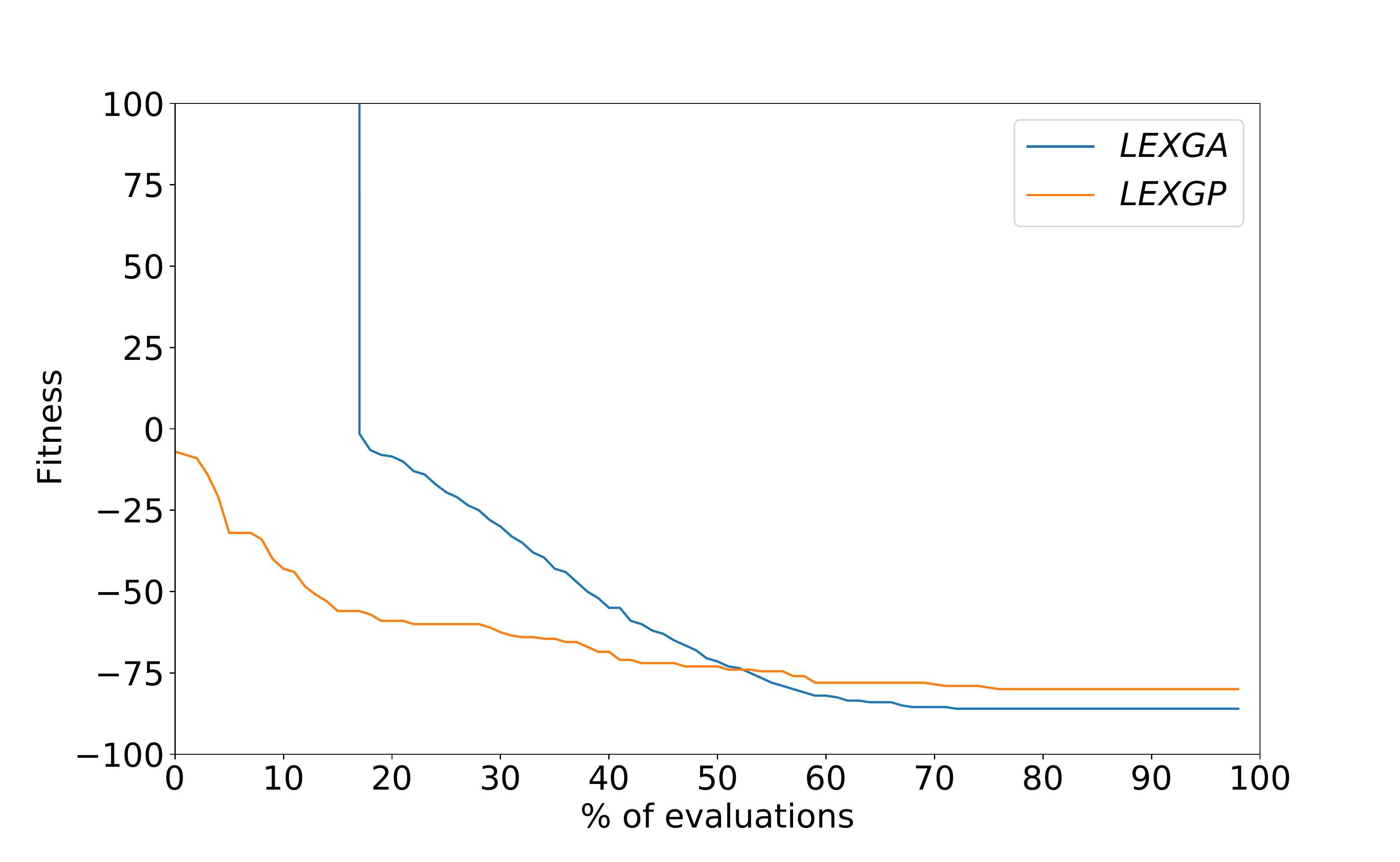} \label{fig:lex_13}} \\
	\subfloat[d=14]{\includegraphics[scale=0.18]{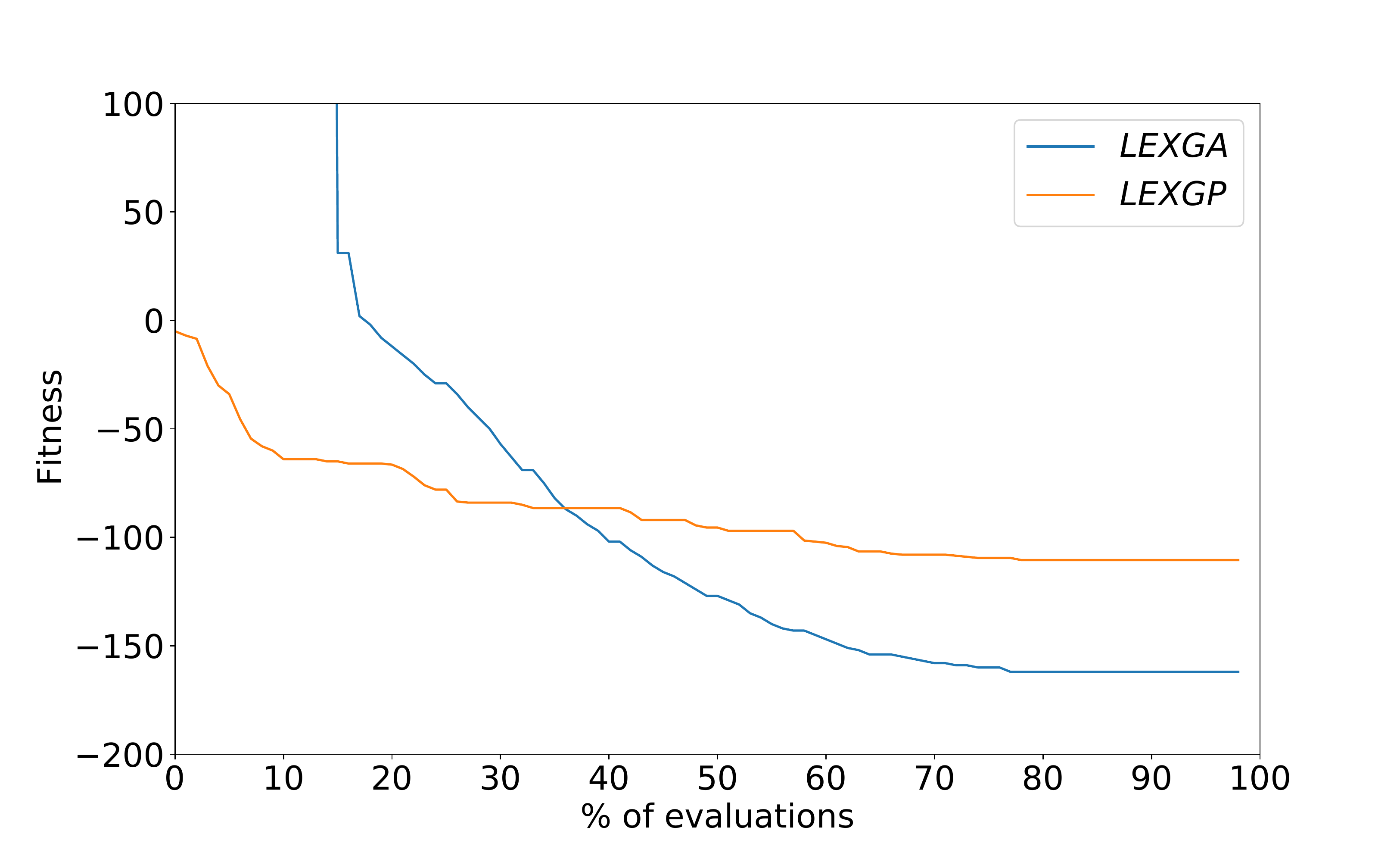} \label{fig:lex_14}} 
	\subfloat[d=15]{\includegraphics[scale=0.18]{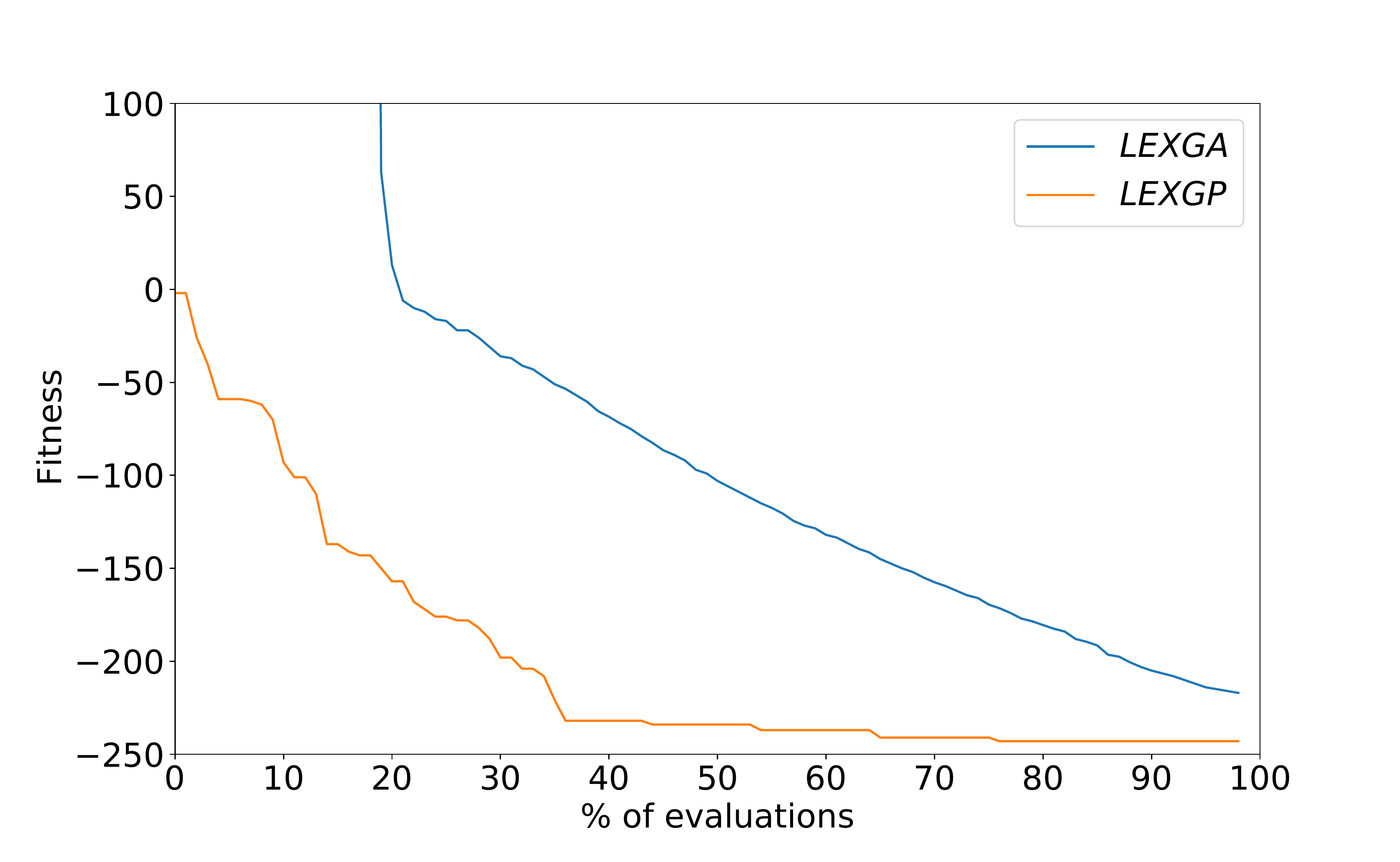} \label{fig:lex_15}} 
	\caption{Lexicographical optimization convergence plots.}
	\label{fig:convergence_lex}
\end{figure}

In Figure~\ref{fig:convergence_lex}, we depict convergence plots for the lexicographic optimization approach.
As before, we depict the median of the best fitness value obtained over all experimental runs. Recall that in this case, the optimization objective is the minimization of $fit_2$, where the compatibility objective is first minimized to get a reversible rule, and then the opposite of $obj_2$ is minimized in order to maximize the Hamming weight of the generating function. Notice also that we cut off the fitness values larger than 100 as GA starts with very large fitness values while GP starts with values close to 0, making the final differences between GA and GP not noticeable. 
Considering GA, we observe around the same percentage of the evaluation required to reach fitness 0 as for the single-objective case. Afterward, it manages to optimize further the Hamming weight making the results comparable with GP. On the other hand, GP again starts with solutions around 0 and slowly improves the fitness value by maximizing the Hamming weight. For sizes up to 14, GA manages to find better final solutions than GP, where the difference is especially noticeable for $d=14$. Interestingly, for $d=15$, GP finds better final solutions but shows no improvement after around 40\% of the fitness budget is used. On the other hand, GA improves the fitness values consistently throughout the evolution process, indicating that with more evaluations, GA could probably reach the performance level of GP.

\subsection{Diversity Analysis}
\label{sec:div}

In Table~\ref{tab:hw-comp}, we provide various diversity metrics for the results obtained for all considered problem instances and algorithms, excluding MOEA. The reason is that we used the multi-objective optimization approach to investigate a different research question not related to the diversity of the solutions. In particular, for each of the four algorithms (SOGA, SOGP, LEXGA, and LEXGP) and diameter $7 \le d \le 15$, we report the number of unique Hamming weights found (UHW), the minimum and maximum Hamming weights observed (respectively mHW and MHW), and the number of unique solutions found (USol). The numbers in bold are the highest values across all methods for each considered diversity metric and diameter.
\begin{table}[t]
	\caption{Diversity metrics for the solutions produced by all optimization methods (excluding MOEA) over all considered diameters.}
	\centering
	\begin{tabular}{cp{0.8cm}ccccccccc}
		\toprule
		\multirow{2}{*}{Algorithm} & \multirow{2}{*}{Metric} & \multicolumn{9}{c}{$d$} \\
		&                         & 7  & 8  & 9  & 10 & 11 & 12 & 13  & 14  & 15  \\
		\midrule
		\midrule
		\multirow{4}{*}{SOGA}      & UHW                     & {\bfseries 5} & 4  & 4  &  6 & 6  & 6  & 6   & 7   & 2   \\
		& mHW                     & 1  & 1  & 1  &  1 & 1  & 1  & 1   & 1   & 1   \\
		& MHW                     & 5  & 4  & 4  &  6 & 6  & 6  & 6   & 8   & 3   \\
		& USol                    & 37 & 46 & {\bfseries 50} & {\bfseries 50} & {\bfseries 50} & {\bfseries 50} & {\bfseries 50}  & 49  & 39  \\
		\midrule
		\multirow{4}{*}{SOGP}      & UHW                     & {\bfseries 5} & {\bfseries 6} & 9  & 8  & 8  & 7  & 9   & 15  & 13  \\
		& mHW                     & 1  & 1  & 1  & 1  & 1  & 1  & 1   & 1   & 1   \\
		& MHW                     & 5  & 8  & 10 & 12 & 16 & 12 & 16  & 19  & 32  \\
		& USol                    & 31 & 34 & 46 & 47 & 49 & 49 & {\bfseries 50}  & {\bfseries 50}  & {\bfseries 50}  \\
		\midrule
		\multirow{4}{*}{LEXGA}     & UHW                     & 2  & 5  & {\bfseries 10} & {\bfseries 16} & {\bfseries 19} & {\bfseries 26} & 33  & 15  & 21  \\
		& mHW                     & {\bfseries 6}  & {\bfseries 8}  & 10 & 15 & {\bfseries 25} & {\bfseries 37} & {\bfseries 65}  & {\bfseries 126} & {\bfseries 206} \\
		& MHW                     & {\bfseries 7}  & {\bfseries 12} & 19 & 30 & 45 & {\bfseries 74} & 116 & {\bfseries 191} & 313 \\
		& USol                    & 23 & 34 & 46 & {\bfseries 50} & {\bfseries 50} & {\bfseries 50} & {\bfseries 50}  & {\bfseries 50}  & {\bfseries 50}  \\
		\midrule
		\multirow{4}{*}{LEXGP}     & UHW                     & 2  & 4  & 5  & 12 & 16 & 19 & {\bfseries 34}  & {\bfseries 28}  & {\bfseries 27}  \\
		& mHW                     & {\bfseries 6}  & {\bfseries 8}  & {\bfseries 16} & {\bfseries 16} & 24 & 32 & 32  & 48  & 95  \\
		& MHW                     & {\bfseries 7}  & {\bfseries 12} & {\bfseries 25} & {\bfseries 30} & {\bfseries 48} & 68 & {\bfseries 128} & 170 & {\bfseries 344} \\
		& USol                    & {\bfseries 45} & {\bfseries 50} & {\bfseries 50} & {\bfseries 50} & {\bfseries 50} & {\bfseries 50} & {\bfseries 50}  & {\bfseries 50}  & 49  \\
		\bottomrule
	\end{tabular}
	\label{tab:hw-comp}
\end{table}

Considering the single-objective algorithms, notice that GP finds more unique Hamming weights, where the differences are small for smaller diameter sizes but become even an order of magnitude larger for greater diameters. The minimal Hamming weight is equal to 1 for both algorithms and all diameter sizes, which is not surprising as single-objective algorithms do not aim to maximize the Hamming weight. The maximal Hamming weights are slightly larger for GP, especially for larger sizes. Again, this is not unexpected as GP finds optimal solutions already in the initial population, while GA required a significant number of evaluations to reach that performance level. On the other hand, if we consider the number of unique solutions found, we observe that GA works better (i.e., it found more diverse solutions). This result is aligned with our previous discussion as GP finds optimal solutions from the beginning, but then it is intuitive that some of those solutions could repeat. Indeed, syntactically different GP trees could map to the same truth table, thus giving rise to the same reversible rule. Going to larger diameter sizes gives good diversity results for GP too, since then, more solutions are optimal.

Next, considering the lexicographic optimization, the number of unique Hamming weights is similar for both GA and GP. The minimal Hamming weight for larger diameters is smaller for GP than GA and similar for smaller diameters. This indicates that the evolution process works better on average for GA, as a larger part of the population exhibits good behavior. For the maximal Hamming weights, we see that GP reaches better results for large diameters, but this is to be expected. Indeed, as GP has solutions with fitness equal to 0 already in the initial population, it can ``use'' the whole evolution process to optimize the Hamming weights. On the other hand, GA requires more than half of evaluations to reach a compatibility score of 0, which means it has much fewer evaluations available to maximize the Hamming weights. Still, the convergence plots indicate that GA progresses well throughout the evolution process. Possibly, adding more evaluations would allow GA to (at least) reach the Hamming weight values obtained with GP.

\begin{figure}[t]
	\centering
	\includegraphics[width=\textwidth]{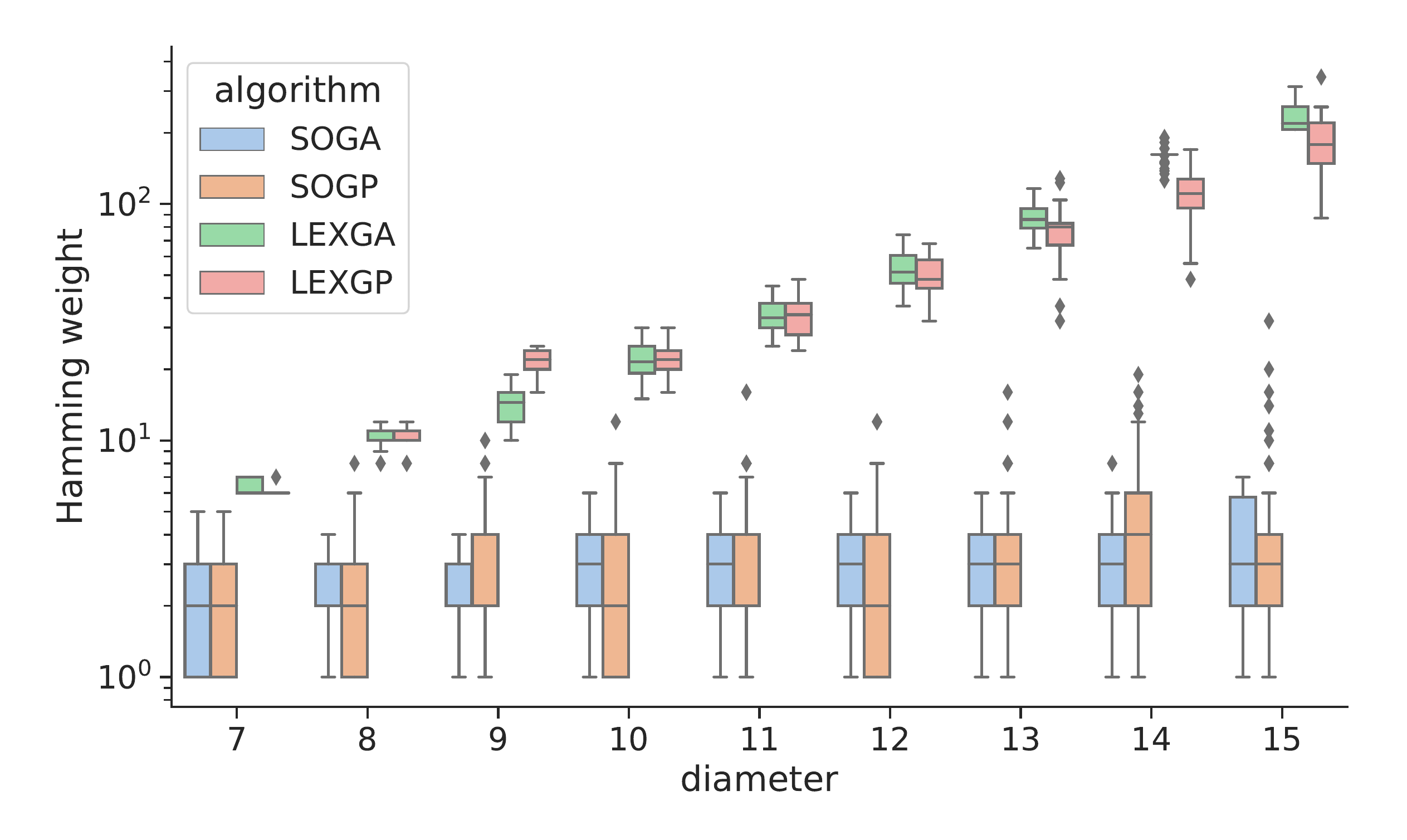}
	\caption{Hamming weight distributions across all compared algorithms and diameters.}
	\label{fig:hw-dist}
\end{figure}

Finally, Figure~\ref{fig:hw-dist} presents the Hamming weight distributions for all considered algorithms and diameter sizes.
First, notice that we can recognize two natural groupings of the distributions, i.e. those related respectively to single-objective optimization and lexicographic optimization. Since in the single-objective optimization, the goal is to reach a fitness value of 0 (i.e., we do not try to maximize the Hamming weight), we can observe that both GA and GP perform similarly and the increase in the Hamming weight value happens only due to a larger diameter (and thus, problem instance). On the other hand, GP performs better on smaller sizes for the lexicographic optimization scenario, which is expected as the initial population already reaches fitness equal to 0, and the obtainable Hamming weight values are relatively close to 0. Considering larger diameters, GA shows slightly better behavior on average. Still, considering the extreme values, we can notice that GP performs better for sizes 13 and 15. Again, this is not surprising as GP has a better ``starting position'', so a greater portion of the evolution process can be used to maximize the Hamming weight. We believe adding more evaluations would resolve this problem and make GA a better performing algorithm, considering the best-obtained values.

\section{Discussion}
\label{sec:disc}
We now discuss the results obtained from our experimental evaluation  applied on Problem~\ref{prob:stat} concerning the three research questions stated in Section~\ref{subsec:rq}.

Concerning RQ1, our experiments in the single-objective optimization scenario further confirm the findings we reported in~\cite{10.1007/978-3-030-44094-7_8}. Indeed, despite the exiguous number of conserved landscape rules compared to the huge size of the search space, both GA and GP always converge to an optimal solution. Moreover, notice that we are dealing with an even smaller optimal set in our current setting than the one adopted in~\cite{10.1007/978-3-030-44094-7_8}, since we set $\omega=3$ instead of $\omega= \lfloor d-1 \rfloor /2$. Nevertheless, this does not seem to pose any significant problem for the considered evolutionary algorithms. There is, however, an important distinction to observe on this statement: while the difficulty for GA to find a reversible rule increases as the diameter gets bigger, GP almost always finds an optimal solution already in the initial population, without even needing to start the evolution process. As mentioned in Section~\ref{subsec:so}, the likely reason for this substantial difference in performances lies in the underlying genotype representations. Arguably, the chances of guessing at random a bitstring of length $2^{d-1}$ that maps to a conserved landscape rule of diameter $d$ are quite low due to the very small number of such rules observed in our exhaustive search experiments. Since the GA population is initialized exactly in this way, it is thus very unlikely that the initial population will already include an optimal individual.
Moreover, a random bitstring will likely have the Hamming weight close to half of its length, or equivalently it will be close to being balanced. As we remarked in Section~\ref{subsec:mo}, balanced bitstrings occur on the top right limit of the Pareto front. Hence they always have the highest possible value concerning the compatibility fitness that one seeks to minimize.

Contrarily, the maximum depth allowed for the trees evolved by GP is linear in the diameter of the local rule, so it is much smaller than the length of the corresponding truth table, which is instead exponential in the diameter. Consequently, it seems reasonable that a random GP tree will map to a truth table with a small Hamming weight. A further explanation of this phenomenon is that in our experiments, we did not use the XOR and XNOR since they were filtered out during the tuning phase. This reduces the probability that the truth table obtained from the evaluation of a GP tree will be balanced, and thus that it will have a large Hamming weight.

Regarding RQ2, our results obtained with the lexicographic optimization approach are also in line with our findings presented in~\cite{10.1007/978-3-030-44094-7_8}. Regardless of the fact that LEXGA and LEXGP were able to find higher Hamming weights than in our previous experiments, they are nonetheless too low to be of any use for cryptographic applications. In particular, we can say something more precise in this respect: as mentioned in Section~\ref{sec:intro}, the Hamming weight is a good proxy for the nonlinearity of a Boolean function, which is a measure of its distance from the set of affine functions. Ideally, Boolean functions of $d$ variables used in stream and block ciphers should have a nonlinearity as high as possible, in the order of $2^{d-1}$ (we refer the reader to~\cite{carlet21} for the reason why this is the case). Moreover, Cusick~\cite{cusick17} showed that the nonlinearity of a $d$-variable Boolean function \emph{coincides} with its Hamming weight if the latter is sufficiently small, i.e., if it is less than $2^{d-2}$. In our case, all generating functions evolved by GA and GP have a Hamming weight which is significantly below $2^{d-2}$, so their nonlinearity corresponds to their weight. Therefore, our results rule out the possibility of using conserved landscape CA in the design of symmetric ciphers components such as filter functions or S-boxes.

Finally, concerning RQ3, the results obtained by our multi-objective optimization experiments further corroborate our previous findings in~\cite{10.1007/978-3-030-44094-7_8}. In particular, also in the case of a fixed offset $\omega$ far from the center of the neighborhood, the Pareto fronts approximated by MOEA show that there is a clear trade-off between the reversibility of a marker CA rule under the conserved landscape definition and its Hamming weight. Moreover, the shapes of the fronts are quite similar to those obtained in~\cite{10.1007/978-3-030-44094-7_8} where the offset was placed at the center. This further suggests that the relationship between the compatibility objective function and the Hamming weight is independent of the cell's position that gets updated in the neighborhood.

\section{Conclusions and Future Works}
\label{sec:conclusions}

In this paper, we considered the search of locally invertible cellular automata defined by conserved landscape rules as a combinatorial optimization problem, using GA and GP to solve it. We based our experimental investigation around three research questions stemming from exhaustive search experiments. To investigate them, we adopted three optimization approaches -- a single-objective, a multi-objective, and a lexicographic optimization approach. After performing a thorough parameter tuning phase, we evaluated the spaces of marker CA rules with diameters between $7$ and $15$, therefore expanding the experiments presented in~\cite{10.1007/978-3-030-44094-7_8} with three additional problem instances. In general, the results obtained from this new set of experiments corroborate the findings of our previous work. In particular, in this new set of experiments, the main new finding is that we fixed the rule offset $\omega$ to $3$ for all problem instances instead of setting it at the center of the neighborhood.
Contrary to our initial assumption, where we hypothesized that this choice would make the optimization problem harder, it turned out to be simpler, especially in the GP case. On the other hand, similar trends of increasing difficulty were observed for GA, although with smaller magnitudes than in the results presented in~\cite{10.1007/978-3-030-44094-7_8}. As argued in Section~\ref{sec:disc}, this difference is most likely caused by the different genotype representations used by GA and GP. Further, the Pareto fronts obtained through our multi-objective optimization experiments not only confirm that the closer a marker CA rule is to be of the conserved landscape type, the lower its Hamming weight must be, but also the converse. Balanced generating functions, which have maximal Hamming weight, are also the farthest possible from inducing a reversible rule. This gave us an additional insight for the reason that GP finds an optimal solution already in the initial population, since the maximum depth enforced on the GP trees is sufficiently small that the resulting truth table will likely have a small Hamming weight.

Several avenues for future research remain to be explored on this subject. Regarding the first research question, the fact that the number of fitness evaluations required for GA to find a conserved landscape rule increases exponentially in the diameter seems to indicate that the difficulty of Problem~\ref{prob:stat} can be easily tuned for optimization algorithms with a bitstring-based representation. This could have in turn potential interesting applications for benchmark purposes. Further, it would be interesting to study this problem from the perspective of runtime analysis. Possibly, one could derive upper bounds on the number of fitness evaluations necessary for a simple evolutionary algorithm to converge on a conserved landscape rule. Likewise, although optimizing only the reversibility property is a trivial problem for GP, it could still be interesting to formally investigate from a theoretical point of view what is the probability of guessing a tree at random that maps to an optimal solution.

For the second research question, our new findings corroborate that the utility of conserved landscape CA for cryptography and reversible computing is quite limited since their Hamming weights are too low concerning the truth table size of their generating functions. Nonetheless, as remarked in Section~\ref{subsec:mark-ca}, one can easily relax the definition of conserved landscape rules by allowing partial overlapping of the landscapes and obtain a larger class of reversible CA with more complex behaviors. A possible idea worth exploring in this direction would be to adapt the fitness function $fit_1$ to allow for this partial overlapping and use GP to investigate the Hamming weights of the resulting reversible CA, in particular with the lexicographic optimization method that proved to be the best performing one.

Finally, for the third research question, as discussed above the Pareto fronts approximated by MOEA showed a clear trade-off between the reversibility of marker CA rules and the Hamming weights of their generating functions. As far as we know, there are no results in the CA literature addressing this aspect of conserved landscape CA. It would thus be interesting to exploit our experimental observation for formally proving an upper bound on the Hamming weight that a conserved landscape CA can achieve.

\subsection*{Source Code and Experimental Data}
The source code and the experimental data related to this paper are available at \url{https://github.com/rymoah/EvoRevCA}.

\bibliographystyle{abbrv}
\bibliography{bibliography}

\end{document}